\documentclass[a4paper,fleqn]{cas-sc}
\usepackage[authoryear]{natbib}
\usepackage{appendix}
\usepackage{threeparttable}
\usepackage{xurl}
\ExplSyntaxOn
\cs_set:Npn \printorcid {}
\cs_set:Npn \__first_footerline: {\small\sffamily Rui~Hua~et~al.:~\textit{Preprint}}
\ExplSyntaxOff
\begin{document}
\let\WriteBookmarks\relax
\def\floatpagepagefraction{1}
\def\textpagefraction{.001}
\shorttitle{Lingdan-V2: Reasoning models for traditional Chinese medicine}
\shortauthors{Rui Hua et al.}
\title[mode=title]{Lingdan-V2: Large language models for clinically grounded reasoning in traditional Chinese medicine}
\author[1]{Rui Hua}
\author[2,3]{Yu Wei}
\author[1]{Ziri Jin}
\author[1]{Zhuo Sun}
\author[1]{Kai Chang}
\author[1]{Jianan Xia}
\author[4,5]{Zixin Shu}
\author[4,5]{Xiaodong Li}
\author[6]{Dongmei Jia}
\author[7]{Fei Dong}
\author[8]{Runshun Zhang}
\author[1]{Jian Yu}
\author[9]{Hao Xu}
\author[10]{Haibin Yu}
\author[10]{Jiansheng Li}
\cormark[1]
\ead{li_js8@163.com}
\author[11]{Baoyan Liu}
\cormark[1]
\ead{liuby5505@139.com}
\author[2,3]{Wenjia Wang}
\cormark[1]
\ead{tsl-wangwenjia@tasly.com}
\author[1]{Xuezhong Zhou}
\cormark[1]
\ead{xzzhou@bjtu.edu.cn}

\affiliation[1]{organization={School of Computer Science \& Technology, Beijing Jiaotong University},
            city={Beijing},
            postcode={100044},
            country={China}}

\affiliation[2]{organization={Tianjin Tasly Digital Chinese Medicine Technology Co., Ltd.},
            city={Tianjin},
            postcode={300410},
            country={China}}

\affiliation[3]{organization={Tasly Biopharmaceuticals Co., Ltd.},
            city={Tianjin},
            postcode={300410},
            country={China}}

\affiliation[4]{organization={Institute of Liver Diseases, Hubei Key Laboratory of the Theory and Application Research of Liver and Kidney in Traditional Chinese Medicine, Hubei Provincial Hospital of Traditional Chinese Medicine},
            city={Wuhan},
            postcode={430061},
            country={China}}

\affiliation[5]{organization={Affiliated Hospital of Hubei University of Chinese Medicine},
            city={Wuhan},
            postcode={430061},
            country={China}}

\affiliation[6]{organization={Da'an Health Technology (Beijing) Co., Ltd.},
            city={Beijing},
            postcode={100020},
            country={China}}

\affiliation[7]{organization={Institute of Chinese Medicine Epidemic Disease, Beijing University of Chinese Medicine},
            city={Beijing},
            postcode={102488},
            country={China}}

\affiliation[8]{organization={Department of Gastroenterology, Guang'anmen Hospital, China Academy of Chinese Medical Sciences},
            city={Beijing},
            postcode={100053},
            country={China}}

\affiliation[9]{organization={Cardiovascular Diseases Center, National Clinical Research Center for Chinese Medicine Cardiology, Xiyuan Hospital, China Academy of Chinese Medical Sciences},
            city={Beijing},
            postcode={100091},
            country={China}}

\affiliation[10]{organization={The First Affiliated Hospital, Henan University of Chinese Medicine},
            city={Zhengzhou},
            postcode={450000},
            country={China}}

\affiliation[11]{organization={China Academy of Chinese Medical Sciences},
            city={Beijing},
            postcode={100700},
            country={China}}

\cortext[1]{Corresponding authors: Jiansheng Li, PhD; Baoyan Liu, PhD; Wenjia Wang, PhD; Xuezhong Zhou, PhD.}

\begin{abstract}
Large language models have advanced medical reasoning, but clinically grounded reasoning in traditional Chinese medicine (TCM) remains challenging because prescription decisions require integrating domain knowledge, syndrome differentiation, treatment principles, herbal prescription composition, and dosage determination. We develop a TCM language model family comprising models with 4 billion, 8 billion, and 14 billion parameters. The artificial intelligence contribution is a staged clinical alignment framework integrating large scale continued pretraining on TCM and clinical records, complementary reasoning supervision, and prescription-centered reinforcement learning (RL). We further formulate herbal prescription generation as a structured RL problem with prescription-level feedback, enabling direct optimization of clinical decision outputs beyond knowledge reproduction and generic instruction following. The engineering application is TCM clinical decision support, particularly herbal prescription recommendation. We evaluate the models on a public benchmark containing 25,620 expert-verified items across 13 TCM knowledge and clinical reasoning tasks and on a private retrospective dataset of 4,348 spleen and stomach disease prescription cases. The 14 billion parameter reinforcement aligned model achieves a prescription generation F1 of 37.0 on the public benchmark, compared with 32.3 for DeepSeek-R1 with 671 billion total parameters, and a herb-level F1 of 52.8 on the clinical dataset with improved dosage agreement. Reinforcement post-training consistently improves prescription recommendation across all scales, while physician-source analysis shows that expert heterogeneity affects reward stability. These findings demonstrate a scalable pathway from domain grounding and reasoning supervision to clinically oriented prescription decision support.
\end{abstract}
\begin{keywords}
Traditional Chinese medicine \sep
Large language models \sep
Reinforcement learning \sep
Clinical reasoning \sep
Prescription recommendation
\end{keywords}
\maketitle
\section{Introduction}

Large language models (LLMs) have transformed natural language processing from task-specific systems into general-purpose models capable of language understanding, reasoning, and interactive decision support, supported by transformer architectures and large-scale pretraining \cite{vaswani2017attention,yang2025qwen3}. In medicine, their application has expanded from knowledge-intensive question answering to diagnostic assistance, clinical communication, and broader decision support \cite{singhal2023large}. Evaluation resources such as GPQA \cite{rein2024gpqa}, MedHELM \cite{bedi2025medhelm}, and Humanity's Next Medical Exam \cite{gallifant2025humanity} have further extended assessment from biomedical knowledge to difficult reasoning and clinical judgment. Clinically useful medical LLMs must therefore move beyond factual recall and fluent generation to integrate heterogeneous patient information, support context-sensitive reasoning, and produce responses that are clinically coherent and appropriately constrained \cite{liu2025application,pengStudyGenerativeLarge2023,tordjman2025comparative}. These requirements are particularly demanding in traditional Chinese medicine (TCM), where clinical reasoning links symptom descriptions, tongue and pulse findings, syndrome differentiation, treatment principles, prescription composition, and dosage decisions within a unified diagnostic and therapeutic process. Existing general medical models and evaluation frameworks provide only limited coverage of this Chinese-language and TCM-specific reasoning setting.

Chinese medical language models and benchmarks have begun to address the linguistic and clinical specificity of Chinese healthcare. CBLUE \cite{zhang2022cblue}, CMB \cite{wang2024cmb}, CMExam \cite{liu2023benchmarking}, and MedBench \cite{cai2024medbench} evaluate Chinese biomedical language understanding, medical knowledge, and examination-oriented reasoning. Models such as HuatuoGPT \cite{zhang2023huatuogpt}, Zhongjing \cite{yang2024zhongjing}, and Baichuan-M2 \cite{dou2025baichuan} further extend general Chinese language models toward medical dialogue and consultation.
TCM-specific research has increasingly focused on domain pretraining, instruction tuning, syndrome differentiation, and clinical task evaluation. Representative models include Tianyi \cite{liu2025tianyi}, TCMChat \cite{dai2024tcmchat}, Qibo \cite{jia2025qibo}, and BianCang \cite{wei2025biancang}. Dedicated benchmarks such as MTCMB \cite{kong2025mtcmb} and TCMEval-SDT \cite{wang2025tcmeval} have expanded evaluation from factual knowledge to syndrome differentiation and multi-task clinical reasoning. Lingdan-V1 \cite{hua2024lingdan} and LingLan \cite{hua2026linglanmidian} similarly demonstrated the importance of TCM-specific training data and systematic evaluation for diagnosis, treatment, and prescription tasks. Nevertheless, most existing approaches remain centered on domain pretraining and supervised adaptation, while the optimization of clinically structured prescription generation and its relationship with broader TCM capabilities remain insufficiently studied.

TCM presents a distinct data and decision structure that is not fully captured by general medical question answering. Its knowledge is distributed across classical texts, modern textbooks, clinical records, materia medica, patent medicine resources, prescription databases, and structured knowledge bases. 
Resources such as TCMID \cite{huang2018tcmid}, SymMap \cite{wu2019symmap}, HERB \cite{fang2021herb}, and ETCM \cite{xu2019etcm} organize relationships among herbs, ingredients, symptoms, diseases, and molecular targets. 
However, clinical prescription recommendation cannot be reduced to retrieving isolated entities. It requires the model to integrate heterogeneous clinical evidence into syndrome differentiation, treatment principles, herb combinations, and dosage decisions. Previous prescription recommendation studies have also shown the importance of modeling herb compatibility and patient-specific syndrome patterns \cite{dong2021tcmpr}. These characteristics make prescription generation a central test of whether a TCM language model can connect domain knowledge with coherent diagnostic and therapeutic decision making.

This challenge cannot be addressed through domain data scaling alone. Continued pretraining can establish TCM knowledge and clinical language representations, while instruction tuning \cite{weiFinetunedLanguageModels2022,wang2023self,chung2024scaling,zhangInstructionTuningLarge2023} and chain-of-thought supervision \cite{weiChainofThoughtPromptingElicits2023} can improve task following and structured reasoning. However, supervised learning primarily imitates observed responses and does not directly optimize agreement with clinically structured prescription decisions. Reward-based post-training provides a complementary approach. Reinforcement learning from human feedback \cite{ouyangTrainingLanguageModels}, direct preference optimization \cite{rafailov2023direct}, and group relative policy optimization (GRPO) \cite{shao2024deepseekmath,guo2025deepseek} have improved alignment and reasoning in general-domain models, but their application to medical reasoning remains constrained by the limited verifiability of clinical outputs \cite{chen2025towards}. TCM prescription recommendation offers a partially verifiable setting because herb composition and dosage patterns can be normalized and compared with reference prescriptions. This creates an opportunity to use prescription-level feedback to align domain knowledge and reasoning capabilities with clinical decision generation.

To address this problem, we present Lingdan-V2, a TCM language model family developed from Qwen3 backbones \cite{yang2025qwen3} at the 4B, 8B, and 14B parameter scales. Continued pretraining uses 16.62 billion TCM-related tokens, including approximately 10.4 billion tokens of electronic medical records (EMR) text, to establish broad domain knowledge and clinically grounded language representations. The supervised fine-tuning (SFT) stage combines 240,000 TCM reasoning samples distilled from DeepSeek-R1 \cite{guo2025deepseek} with publicly available Chinese reasoning instruction data to support TCM-specific reasoning and general instruction following. Prescription-centered GRPO then uses normalized agreement with reference herbs as the principal reward signal to further align the supervised models with clinical prescription generation. Full-parameter optimization is applied throughout the training pipeline, and the paired Lingdan-SFT and Lingdan-R1 checkpoints at each model scale enable direct analysis of the changes associated with prescription-oriented reinforcement post-training.

We evaluate Lingdan-V2 on the LingLan Benchmark \cite{hua2026linglanmidian}, together with a private retrospective clinical prescription dataset.
The contributions of this study are as follows:
First, we develop a clinically grounded, multi-scale TCM model family that integrates large-scale domain learning with reasoning supervision. 
Second, we formulate prescription generation as a reinforcement-learning task and identify physician heterogeneity as a source of reward instability. 
Third, we evaluate Lingdan-V2 across model scales, benchmark difficulty, and retrospective clinical prescription settings, demonstrating broad TCM performance and consistent prescription gains.
\section{Materials and Methods}

This section describes the development and evaluation of Lingdan-V2. We first introduce the model family and training data, then present data processing, staged optimization and prescription-oriented GRPO, and finally define the public and private evaluation frameworks and the measures used to support reproducibility.

\subsection{Study Design and Model Overview}

\begin{figure}[t]
\centering
\includegraphics[width=\linewidth]{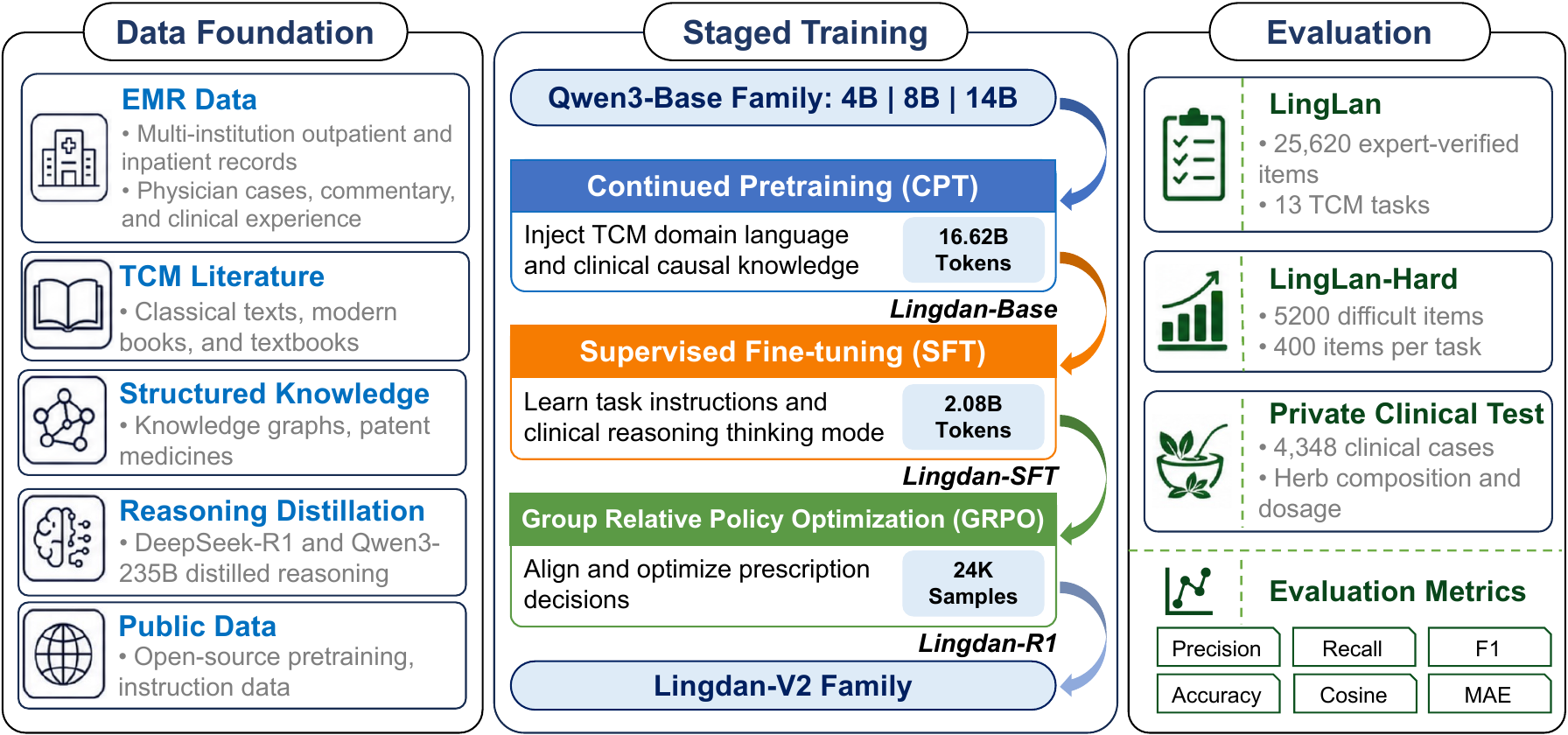}
\caption{Overview of the Lingdan-V2 development and evaluation workflow. Qwen3 backbones at 4B, 8B and 14B scales were trained using the same CPT, SFT and GRPO pipeline. Evaluation was performed using LingLan, LingLan-Hard and a private clinical prescription recommendation test set. MAE, mean absolute error.}
\label{fig:lingdan_overview}
\end{figure}

This study developed and evaluated Lingdan-V2, a family of TCM-oriented large language models derived from Qwen3 foundation models \cite{yang2025qwen3}. 
The family includes 4B, 8B and 14B variants, all of which were trained using the same staged adaptation framework consisting of continued pretraining (CPT), supervised fine-tuning (SFT), and group-relative policy optimization (GRPO). 
The supervised checkpoints are denoted Lingdan-\{scale\}-SFT, and the corresponding GRPO-aligned models are denoted Lingdan-\{scale\}-R1, where \{scale\} represents the model size (4B, 8B or 14B).
The 14B models were used as the primary models for detailed evaluation and analysis, whereas the 4B and 8B variants were included to characterize the impact of model scale. 
Lingdan-14B-SFT and Lingdan-14B-R1 were compared to assess the performance change associated with GRPO alignment after the preceding CPT and SFT stages. 
Figure~\ref{fig:lingdan_overview} summarizes the data foundation, staged training procedure and evaluation settings for Lingdan-V2.

\subsection{Training Data Construction}

Training data were organized according to their roles in the three-stage adaptation framework, including domain adaptation and knowledge enhancement through CPT, task-oriented instruction learning through SFT, and prescription-centered reward optimization through GRPO.
We describe the large-scale corpus used for CPT, the consolidated SFT corpus, and the clinical prescription source pool used for GRPO in sequence.
Publicly available datasets were incorporated to complement private TCM and clinical resources. 
These datasets covered TCM knowledge, medical question answering, clinical dialogue, examination-oriented tasks, instruction following, and general reasoning. The detailed provenance and category of public datasets are summarized in Table~\ref{tab:public_datasets}.

\subsubsection{Continued Pretraining Corpus}

The CPT corpus contained 16.62B tokens after preprocessing and final consolidation. It integrated private TCM and clinical resources with publicly available corpora to adapt the Qwen3 backbone to TCM-specific language patterns, clinical knowledge and general medical contexts. Private resources included longitudinal patient care trajectories, knowledge-graph-derived text, Chinese patent medicine and drug-instruction data, classical TCM texts with modern Chinese translations, modern TCM books with case-commentary material, and Lingdan V1 data \cite{hua2024lingdan}. Public resources consisted of open-source pretraining corpora, medical question-answering resources, and general-domain text corpora. The composition of the final CPT corpus is summarized in Table~\ref{tab:cpt_data}.

\begin{table}[!htbp]
    \centering
    \small
    \caption{Continued pretraining corpus used for Lingdan-V2 model development after final consolidation. Token counts are reported in billions.}
    \label{tab:cpt_data}
    \begin{tabular}{lrrl}
    \toprule
    Data category & Tokens (B) & Share (\%) & Domain \\
    \midrule
    Longitudinal patient care trajectories corpus & 10.42 & 62.72 & TCM/EMR \\
    Clinical diagnosis and physician-assistance data & 0.50 & 2.98 & TCM/EMR \\
    Entity extraction data & 0.44 & 2.66 & TCM \\
    Lingdan V1 data & 0.32 & 1.90 & TCM \\
    Classical texts with translations & 0.29 & 1.75 & TCM \\
    Knowledge graph and patent medicine question answering & 0.18 & 1.11 & TCM \\
    Modern TCM books & 0.09 & 0.57 & TCM \\
    Open-source public data & 3.65 & 21.97 & General/medical \\
    Benchmark-derived training data & 0.72 & 4.34 & General/medical \\
    \midrule
    Total & 16.62 & 100.00 & -- \\
    \bottomrule
    \end{tabular}
\end{table}

\subsubsection{Supervised Fine-tuning Data}

The SFT corpus was constructed to support both general instruction following and TCM-specific clinical reasoning. The source pool comprised open-source instruction data, benchmark-oriented task distillation, entity extraction annotations, clinical diagnosis completion, TCM knowledge question answering, Chinese patent medicine tasks, physician-style clinical reasoning, physician-synthesized cases, physician commentary generation, spleen-stomach prescription generation, TCM licensing questions, and Qwen3-235B distillation \cite{yang2025qwen3}. Benchmark-oriented distillation refers to the construction of training examples following the task formats represented in public benchmarks. A spleen-stomach prescription generation subset containing 43,469 samples and 11.46M tokens was retained for subsequent prescription-centered GRPO training. All source datasets were converted into standardized training files before merging. 
After deduplication, format cleaning, length filtering, quality screening, sampling and final consolidation, the resulting SFT corpus contained 7.73 million samples and 2.08B tokens. 
Its final composition, grouped into mutually exclusive categories at the training-file level, is summarized in Table~\ref{tab:sft_source_data}.

\begin{table}[!htbp]
    \centering
    \caption{Final composition of the consolidated SFT corpus. Categories are mutually exclusive at the final training-file level; token counts are rounded to two decimals.}
    \label{tab:sft_source_data}
    \begin{tabular}{lrr}
    \toprule
    Training component & Samples & Tokens (M) \\
    \midrule
    Direct-answer instruction data & 6,879,359 & 1,238.49 \\
    Reasoning-format SFT data & 200,529 & 216.24 \\
    Instruction distillation with reasoning traces & 46,221 & 70.33 \\
    Instruction data without reasoning traces & 75,741 & 26.98 \\
    TCM multi-turn dialogue data & 18,468 & 8.45 \\
    Physician-synthesized cases & 262,987 & 155.18 \\
    Physician commentary generation & 26,677 & 32.42 \\
    Qwen3-235B distillation & 220,000 & 336.75 \\
    \midrule
    Total & 7,729,982 & 2,084.85 \\
    \bottomrule
    \end{tabular}
\end{table}

\subsubsection{GRPO Training Data}

The GRPO source pool contained 24,212 clinical prescription-generation prompts after filtering, comprising 9,849 augmented physician-style cases and 14,363 cleaned spleen-stomach disease prescription cases. The physician-style cases were derived from an original corpus of approximately 20,000 physician-authored cases contributed by more than 40 physicians. After quality-based selection, cases from five physicians were retained as the basis for augmentation, resulting in 9,849 physician-style training cases after data augmentation and quality screening. The spleen-stomach disease prescription cases were independently collected and cleaned for prescription-generation optimization. During GRPO training, 25,000 prompts were sampled from the physician-style source and 15,000 prompts from the spleen-stomach source, yielding a 40,000-prompt training stream. Because the requested sampling sizes exceeded the available source sizes, source records were repeatedly sampled during training stream construction.

\subsection{Data Processing and Quality Control}

Training data from heterogeneous sources required standardized preprocessing before model development. Data processing consisted of source-specific normalization followed by a shared quality-control pipeline. Source-specific procedures were applied to transform different data types, including clinical records, TCM literature, structured knowledge resources and instruction datasets, into standardized model-ready representations. All processed data subsequently underwent common deduplication, formatting validation and quality screening before stage-specific consolidation.

\subsubsection{Clinical Data Standardization and De-identification}

\begin{figure}[!htbp]
    \centering
    \includegraphics[width=\linewidth]{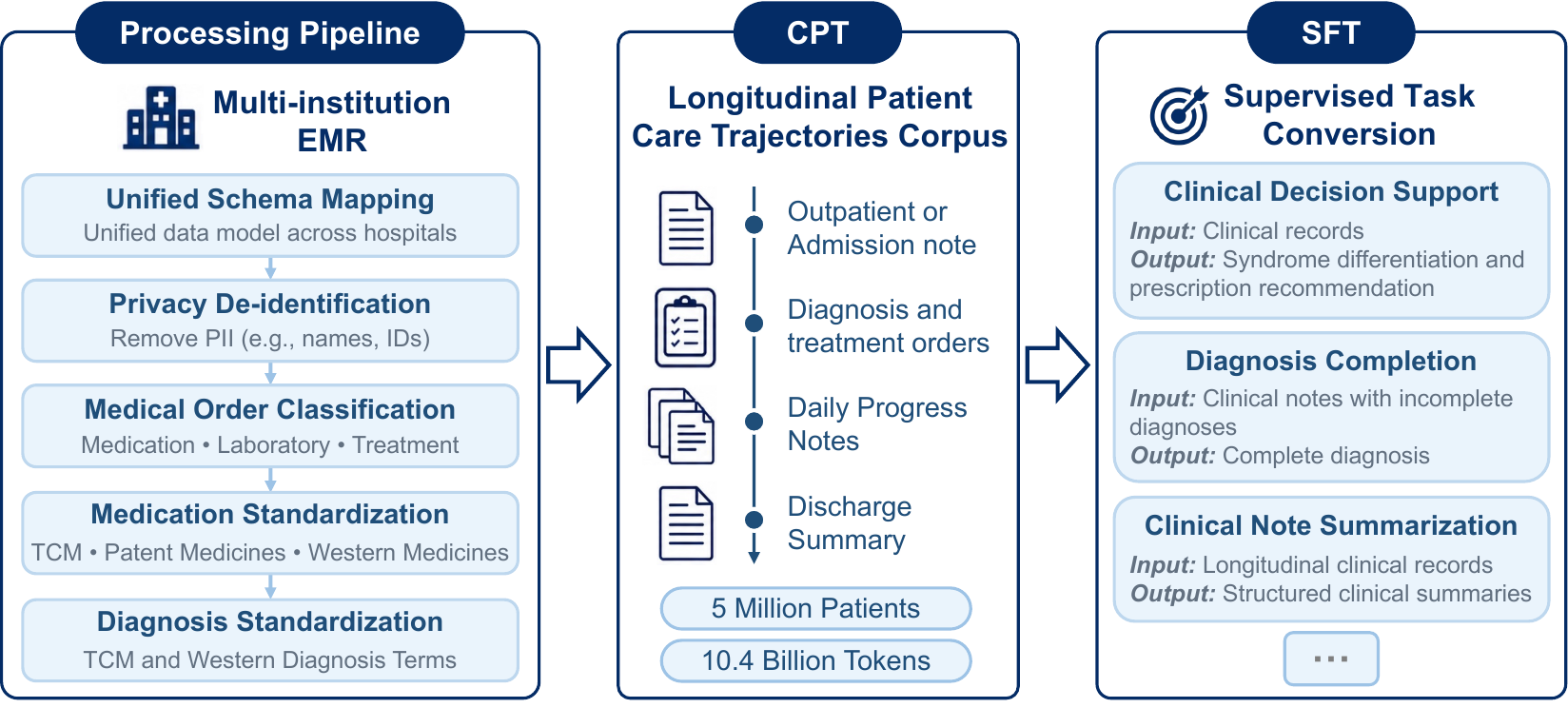}
    \caption{Clinical data processing and task construction. Heterogeneous outpatient and inpatient records are mapped to a common schema, de-identified, and standardized for medical orders, medications and diagnoses. Longitudinal patient trajectories support CPT, whereas task conversion produces SFT data for clinical decision support, diagnosis completion and clinical-note summarization. PII, personally identifiable information.}
    \label{fig:clinical_data_process}
\end{figure}

Private clinical data required additional preprocessing because of their heterogeneous structures and sensitive nature. Clinical records were derived from outpatient and inpatient information systems with institution-specific database schemas. Hospital-specific schemas were mapped into a unified data structure, and medical terminology, including drug names and diagnosis names, was normalized. Direct identifiers and potentially identifying information were removed or masked before downstream processing. Clinical records were organized into task-relevant categories, including demographics, diagnoses, medications, laboratory results, clinical notes and treatment records. Structured laboratory indicators and clinical text were integrated with longitudinal patient trajectories to generate model-ready clinical sequences for continued pretraining, supervised fine-tuning and reinforcement-learning data construction. Figure~\ref{fig:clinical_data_process} summarizes the clinical data processing pipeline and the conversion of longitudinal patient trajectories into CPT and SFT data.

\subsubsection{Cross-stage Data Quality Control}

All training corpora underwent a shared quality-control pipeline before stage-specific consolidation. We applied MinHash-based deduplication, length filtering, quality screening and final consolidation. Reasoning-format data were additionally cleaned to ensure valid paired \texttt{<think>} and \texttt{</think>} tags, and sliding-window rules were applied to remove abnormal repeated-character patterns. During GRPO preparation, prompts whose completion groups had zero reward standard deviation were dynamically resampled, with a maximum of five attempts, because such groups provide limited relative-preference learning signals.

\subsection{Model Training Procedure}

Lingdan-V2 models were trained using a three-stage adaptation framework consisting of CPT, SFT, and prescription-centered GRPO.
All 4B, 8B and 14B models followed the same sequential training procedure. All stages used full-parameter training with bfloat16 precision and stage 2 of the Zero Redundancy Optimizer (ZeRO-2) optimization. 
Learning rates were adjusted according to model scale, with larger models using smaller learning rates.
During CPT, Qwen3 base models were adapted to the TCM domain using the consolidated corpus described above. 
CPT optimized the standard causal language modeling objective over all tokens in the training sequences, enabling the models to acquire TCM-specific language patterns and clinical knowledge representations. 
This stage produced the Lingdan-Base models. 
The primary 14B model was trained on 32 NVIDIA A100 80GB GPUs across four nodes for approximately 120 hours.
During SFT, the CPT-adapted models were optimized using instruction-response supervision. 
The SFT objective was computed on assistant responses, while instruction tokens were masked during optimization. 
Reasoning-format SFT samples containing explicit \texttt{<think>} and \texttt{</think>} traces were included to initialize structured reasoning behaviors before GRPO alignment. 
This stage produced the Lingdan-SFT models.
Starting from the reasoning-capable SFT checkpoints, prescription-centered GRPO was applied to further align model outputs with the clinical prescription-generation objective. 
This stage optimized policy updates using prescription-level feedback and produced the Lingdan-R1 models, with Lingdan-14B-R1 used as the primary model evaluated in this study.

\subsection{GRPO Objective and Reward Design}

The GRPO stage optimized prescription generation using groups of eight sampled completions and a composite reward. Policy updates used group-relative advantages and a clipped objective without Kullback--Leibler (KL) divergence regularization. The optimization objective, clipping parameters, and zero-variance resampling procedure are detailed in Appendix~\ref{app:grpo_details}.

\subsubsection{Prescription-centered Composite Reward}

The composite reward was defined as

\begin{equation}
r = 0.9r_{\mathrm{pres}} + 0.1r_{\mathrm{format}} + 0.1r_{\mathrm{length}},
\end{equation}

where $r_{\mathrm{pres}}$, $r_{\mathrm{format}}$ and $r_{\mathrm{length}}$ denote the prescription agreement reward, format reward and length penalty, respectively.

The format reward was set to one when a completion contained exactly one valid \texttt{<think>} and \texttt{</think>} pair and a parseable prescription in JavaScript Object Notation (JSON) format; otherwise, it was set to zero.

Model outputs were parsed as TCM herbal prescriptions. Herb names were normalized using a predefined herb normalization dictionary $\mathcal{D}_{\mathrm{herb}}$, which contained standardized names, synonym mappings and inclusion-based matching rules. A one-to-one matching procedure was then applied between predicted herbs and reference herbs according to $\mathcal{D}_{\mathrm{herb}}$. Let $m$ denote the number of matched herb pairs, and let $n_{\mathrm{pred}}$ and $n_{\mathrm{ref}}$ denote the numbers of herbs in the generated and reference prescriptions. The prescription reward was calculated as the normalized herb-level F1 score:

\begin{equation}
r_{\mathrm{pres}} =
\frac{2m}{n_{\mathrm{pred}}+n_{\mathrm{ref}}}.
\end{equation}

A soft length penalty discouraged overly long completions; its definition and parameters are provided in Appendix~\ref{app:grpo_details}.

\subsection{Evaluation Framework}

The evaluation framework consisted of benchmark-based evaluation and private clinical prescription recommendation evaluation. The following subsections describe the evaluation datasets and metrics, followed by the shared inference, output-processing and score-aggregation protocols.

\subsubsection{Evaluated Models and Baselines}

The Lingdan-V2 evaluation comprised six checkpoints across the 4B, 8B, and 14B parameter scales. At each scale, the Lingdan-SFT checkpoint was paired with its corresponding Lingdan-R1 checkpoint to characterize performance changes associated with prescription-centered GRPO. Lingdan-14B-R1 was used as the primary model for comparison with external systems, while the 4B and 8B pairs were included to examine the consistency of prescription-related changes across model scales.

The general-purpose and reasoning-oriented comparison set covered a broad range of model sizes, architectures, and post-training strategies. The Qwen3 family references included Qwen3-4B, Qwen3-8B, Qwen3-14B, Qwen3-32B, Qwen3-30B-A3B, Qwen3-235B, and Qwen3-Next-80B-A3B \cite{yang2025qwen3}. These models provided same-family references for assessing the effects of TCM domain adaptation and prescription-oriented post-training. Additional baselines included DeepSeek-R1 \cite{guo2025deepseek}, DeepSeek-V3.1 in thinking mode, GPT-5, GPT-5-mini, GPT-OSS-20B, GPT-OSS-120B \cite{agarwal2025gpt}, and Baichuan-M2-32B \cite{dou2025baichuan}. Together, the comparison set covered both similarly sized models and substantially larger reasoning systems.
TCM-oriented baselines were included to position Lingdan-V2 relative to existing domain-specialized models. The comparison covered Lingdan-13B-TCPM \cite{hua2024lingdan}, TCMChat-7B \cite{dai2024tcmchat}, Baichuan2-13B-Chat \cite{yang2023baichuan}, BianCang-7B and BianCang-14B \cite{wei2025biancang}, and ShizhenGPT-7B and ShizhenGPT-32B \cite{chen2025shizhengpt}. This group represented previous Lingdan development together with TCM models based on continued pretraining, instruction tuning, and domain-specific adaptation.

All included models were evaluated on the full LingLan benchmark and LingLan-Hard under the same inference and scoring protocol. The private spleen-stomach disease evaluation included the six Lingdan-V2 checkpoints, Qwen3-4B, Qwen3-8B, Qwen3-14B, Qwen3-32B, ShizhenGPT-7B, and ShizhenGPT-32B. Qwen3-32B was presented in both LingLan result matrices as a boundary reference between model groups, with both tables reporting results from the same evaluation.

\subsubsection{LingLan Benchmark Evaluation}

LingLan contained 25,620 expert-verified items covering TCM licensing-style examination, foundational TCM knowledge, Chinese patent medicine knowledge, information extraction and diagnostic-therapeutic decision-making tasks \cite{hua2026linglanmidian}. LingLan-Hard was a predefined high-difficulty subset containing 5,200 items, with 400 samples selected from each of 13 tasks. External baselines and Lingdan variants were evaluated using the same prompting, inference and scoring protocols. The compared models are listed in Table~\ref{tab:linglan_large_matrix} and  Table~\ref{tab:linglan_small_matrix}, and cross-task performance was summarized using the Overall Score defined below.

\subsubsection{Private Clinical Prescription Evaluation}

The private spleen-stomach disease evaluation set contained 4,348 de-identified clinical cases. This evaluation followed the same prescription recommendation task formulation and scoring protocol as the prescription recommendation subtask in LingLan. Given clinical case information, models generated herbal prescription compositions and dosages. Herb names were matched using bidirectional string inclusion and evaluated using precision, recall and F1. For dosage evaluation, matched herbs were aligned in a one-to-one manner, while unmatched predicted and reference herbs were assigned a dose of zero before calculating cosine similarity. Dose mean absolute error (MAE) was calculated over matched herb pairs. The private evaluation set was used exclusively to assess model performance in real-world clinical prescription generation.

\subsubsection{Inference and Scoring Protocol}

To ensure fair cross-model comparison, all models were evaluated using the same task prompts, decoding configurations, output requirements and scoring procedures. Each benchmark item was evaluated in a zero-shot setting using a task-specific Chinese user template. All evaluations used a maximum generation length of 8,192 tokens and a temperature of 0.6. Private prescription evaluation followed the same inference and scoring protocol.

Model outputs were processed using task-specific parsing procedures before scoring. Choice tasks were evaluated based on extracted final answers. Fill-in-the-blank responses were evaluated using character-level F1. Information-extraction outputs were evaluated using entity-level matching. Outputs for diagnostic and therapeutic decision-making tasks were evaluated according to predefined matching rules for syndrome, treatment and prescription fields. Prescription outputs were normalized into structured herb-dose pairs for downstream evaluation. Per-task metrics were calculated across all examples in the corresponding evaluation split. To summarize performance across tasks, we computed an Overall Score as the unweighted arithmetic mean of 16 primary metrics. These metrics included licensing-examination accuracy; three fundamental-TCM knowledge metrics; three Chinese patent medicine knowledge metrics; clinical-record and classical-text extraction F1; syndrome F1, treatment F1, prescription F1, dosage cosine similarity; and three diagnostic and therapeutic decision-making accuracies. Dose MAE was excluded from the Overall Score because lower values indicate better performance and therefore are not directly comparable with the percentage-based metrics included in the aggregate score.

\subsection{Reproducibility and Implementation Details}

All model training experiments were conducted using full-parameter training under a distributed training environment. The models were trained and evaluated using standardized configurations described above, including consistent data preprocessing, model templates and evaluation protocols across experiments. The Lingdan-V2 model checkpoints, training configurations and evaluation resources will be released to facilitate reproducibility and further research.

\section{Results}

\begin{figure}[!t]
    \centering
    \includegraphics[width=\linewidth]{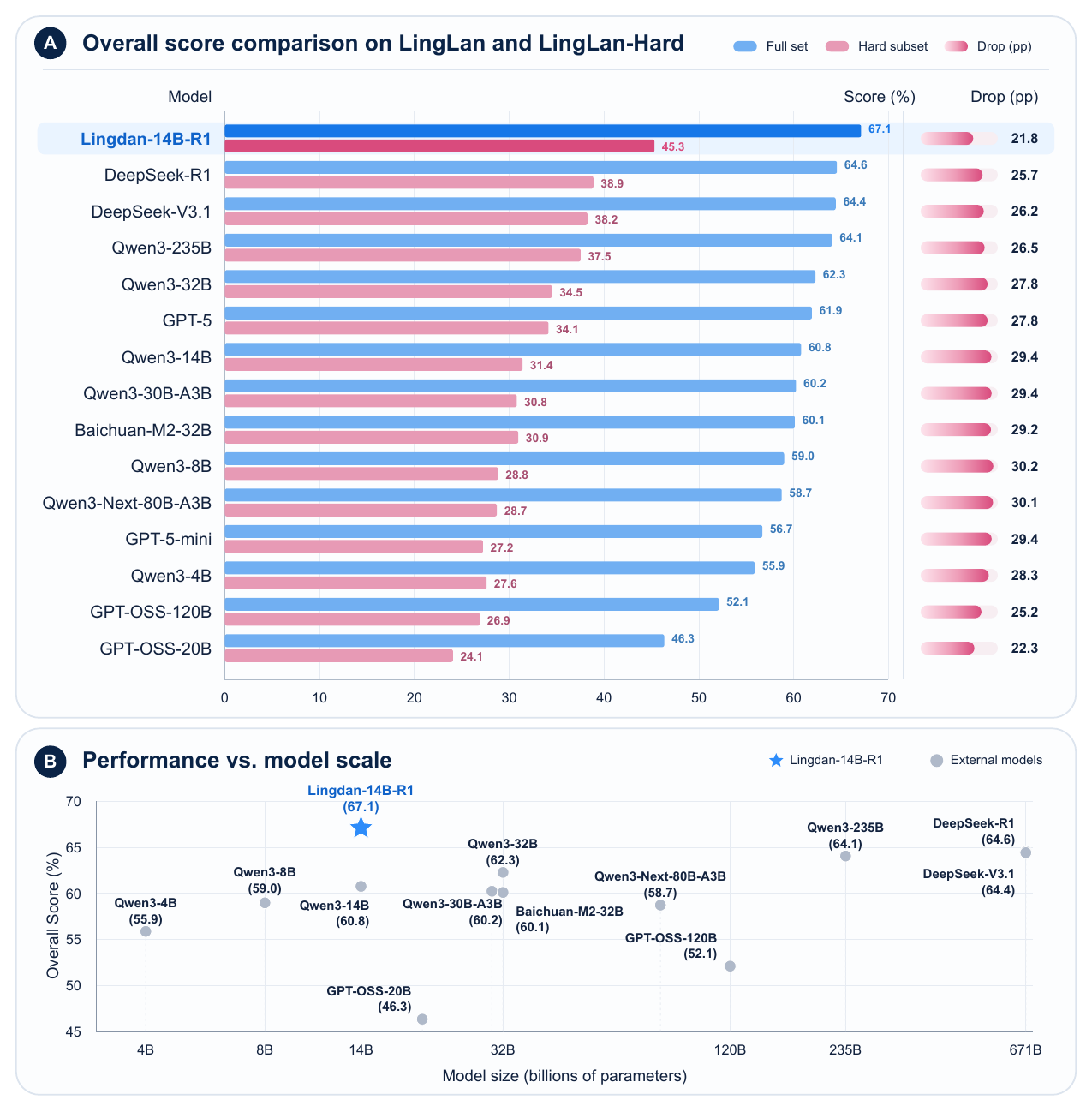}
    \caption{Overall Score comparison on LingLan. Panel A shows Overall Scores on the full benchmark and LingLan-Hard subset for Lingdan-14B-R1 and external models, together with the performance drop on the harder subset. Panel B illustrates Overall Score as a function of model scale, showing the relative position of Lingdan-14B-R1 among models with different parameter sizes. Performance drops are expressed in percentage points (pp).}
    \label{fig:lingdan_overall_comparison}
\end{figure}
    
We evaluated Lingdan-V2 across benchmark-based and clinical prescription-generation settings, focusing on overall benchmark performance, GRPO-associated changes across model scales, and transfer to private clinical prescription cases.

\subsection{Benchmark Performance}

We evaluated Lingdan-14B-R1 on the full LingLan benchmark and its Hard subset. All benchmark metrics except dose MAE are reported on a 0--100 scale. Lingdan-14B-R1 achieved an Overall Score of 67.1 on the full benchmark and 45.3 on LingLan-Hard, compared with 60.8 and 31.4, respectively, for the original Qwen3-14B baseline \cite{yang2025qwen3}. Among the external baselines, DeepSeek-R1 achieved the strongest overall performance, with scores of 64.6 and 38.9 on the full and Hard splits, respectively. Lingdan-14B-R1 therefore outperformed both its same-scale Qwen3 backbone and substantially larger external models. Its margin over the strongest external baseline was 2.5 points on the full benchmark and 6.4 points on LingLan-Hard. Figure~\ref{fig:lingdan_overall_comparison} presents the overall comparison, and complete task-level results are reported in Table~\ref{tab:linglan_large_matrix} and Table~\ref{tab:linglan_small_matrix}.

We next examined knowledge-oriented tasks.
On the TCM Licensing Examination, Lingdan-14B-R1 achieved accuracies of 91.7 and 67.0 on the full and Hard splits, respectively, compared with 95.5 and 81.0 for DeepSeek-V3.1, the strongest external model on this task. 
Across the three primary Fundamental TCM Knowledge metrics, Lingdan-14B-R1 achieved single-choice accuracies of 88.2 and 54.8, multiple-choice F1 scores of 91.1 and 79.4, and cloze F1 scores of 60.2 and 23.9. 
The corresponding strongest external results were 86.1 and 46.5 for DeepSeek-R1, 91.0 and 77.8 for DeepSeek-R1, and 59.9 and 23.5 for Qwen3-235B. 
Across the three primary Chinese Patent Medicine Knowledge metrics, Lingdan-14B-R1 achieved single-choice accuracies of 78.1 and 38.3 and multiple-choice F1 scores of 87.7 and 76.6, exceeding the corresponding DeepSeek-R1 results of 74.3 and 31.0 and 87.0 and 72.4. 
Its cloze F1 scores were 73.4 and 57.4; the strongest external results were 74.4 on the full benchmark for DeepSeek-V3.1 and 57.6 on LingLan-Hard for DeepSeek-R1.
For information extraction, Lingdan-14B-R1 achieved clinical-record extraction F1 scores of 69.6 and 55.5 on the full and Hard splits, respectively, and classical-text extraction F1 scores of 62.6 and 45.6. 
The mean F1 across the two information-extraction tasks was 66.1 on the full benchmark and 50.6 on LingLan-Hard. 
The strongest external-model results for this summary measure were 63.2 on the full benchmark for Qwen3-235B and 46.9 on LingLan-Hard for Qwen3-14B. 
Figure~\ref{fig:linglan_information_extraction} presents the corresponding comparison.

\begin{figure}[!htbp]
    \centering
    \includegraphics[width=\linewidth]{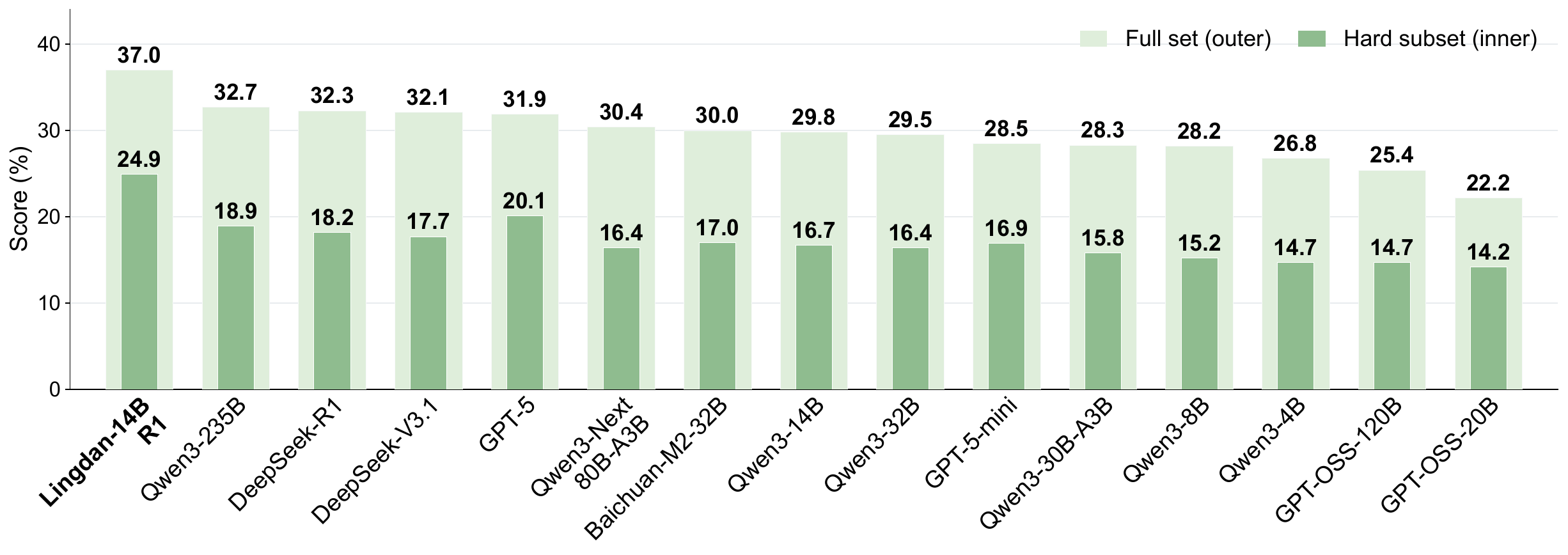}
    \caption{Prescription recommendation on LingLan. Wide light-green bars show prescription-recommendation F1 on the full benchmark, and narrow dark-green bars show F1 on LingLan-Hard.}
    \label{fig:linglan_prescription}
\end{figure}

For diagnostic and therapeutic decision-making, Lingdan-14B-R1 achieved syndrome differentiation F1 scores of 12.1 and 2.4 on the full and Hard splits, compared with the strongest external results of 11.9 for DeepSeek-V3.1 on the full benchmark and 1.6 on LingLan-Hard. 
Its treatment principle F1 scores were 15.5 and 2.1, compared with the strongest external results of 14.8 for DeepSeek-R1 on the full benchmark and 2.8 for GPT-5 on LingLan-Hard. 
The largest margins were observed in prescription-oriented generation, as shown in Figure~\ref{fig:linglan_prescription}. 
Lingdan-14B-R1 achieved prescription-generation F1 scores of 37.0 and 24.9, compared with the strongest external results of 32.7 for Qwen3-235B on the full benchmark and 20.1 for GPT-5 on LingLan-Hard. 
Its dosage cosine similarity scores were 35.4 and 23.7, exceeding the corresponding GPT-5 results of 31.4 and 19.7. 
Lingdan-14B-R1 also led all external models in syndrome, treatment and prescription decision recognition, achieving accuracies of 90.8, 86.9 and 93.4 on the full benchmark and 57.0, 47.0 and 69.3 on LingLan-Hard, respectively.

To position Lingdan-V2 against existing TCM-oriented systems, we separately compared Lingdan-14B-R1 with seven TCM-oriented and related Chinese-language baselines under the same evaluation protocol (Table~\ref{tab:linglan_domain_matrix}). Lingdan-14B-R1 achieved Overall Scores of 67.1 and 45.3 on the full and Hard splits, whereas the strongest scores among the seven comparison models were 59.8 and 36.3. This corresponded to margins of 7.3 and 9.0 points, respectively. The difference was also evident in prescription generation: Lingdan-14B-R1 achieved F1 scores of 37.0 and 24.9, compared with the strongest comparator scores of 27.0 and 16.8.

\newcommand{\modelhead}[2]{\shortstack{\textbf{#1}\\[-1pt]\textbf{#2}}}
\newcommand{\modelheadthree}[3]{\shortstack{\textbf{#1}\\[-1pt]\textbf{#2}\\[-1pt]\textbf{#3}}}

\begin{table}[!t]
\centering
\setlength{\tabcolsep}{2pt}
\renewcommand{\arraystretch}{0.68}
\caption{Complete LingLan task-level results for large and strong external baseline models and the final Lingdan-14B-R1 model. Cells report full-set / Hard-subset scores; bold marks the best value within this table and split, with lower values better for MAE. The final row reports Overall Score, as defined in Methods. Abbreviations: TLE, TCM Licensing Examination; FTK, Fundamental TCM Knowledge; CPMK, Chinese Patent Medicine Knowledge; IE, Information Extraction; DTR, Diagnostic and Therapeutic Reasoning; DTR-F1, DTR with synonym-tolerant character-level matching; DR, Decision Recognition.}
\label{tab:linglan_large_matrix}
\resizebox{\linewidth}{!}{%
\begin{tabular}{lllccccccccc}
\toprule
\textbf{Dom.} & \textbf{Subtask} & \textbf{Metric} & \modelhead{DeepSeek-R1}{671B} & \modelhead{DeepSeek-V3.1}{671B} & \modelhead{GPT-5}{Undisclosed} & \modelhead{Qwen3}{235B} & \modelhead{Qwen3}{32B} & \modelhead{Baichuan-M2}{32B} & \modelhead{Qwen3-Next}{80B-A3B} & \modelhead{GPT-OSS}{120B} & \modelhead{Lingdan-14B}{R1} \\
\midrule
TLE & Comprehensive & Accuracy & 95.0 / 78.3 & \textbf{95.5} / \textbf{81.0} & 87.6 / 52.3 & 95.0 / 77.8 & 91.2 / 62.3 & 89.8 / 58.3 & 93.8 / 73.0 & 58.1 / 14.5 & 91.7 / 67.0 \\
\midrule
FTK & Single-choice & Accuracy & 86.1 / 46.5 & 86.1 / 45.3 & 85.4 / 44.5 & 85.0 / 45.3 & 86.1 / 44.3 & 83.4 / 35.8 & 82.1 / 33.3 & 70.4 / 28.2 & \textbf{88.2} / \textbf{54.8} \\
FTK & Multiple-choice & Accuracy & \textbf{64.8} / \textbf{31.5} & 62.4 / 28.0 & 53.9 / 19.3 & 61.3 / 25.0 & 56.3 / 23.3 & 48.6 / 14.0 & 52.6 / 14.8 & 35.3 / 9.8 & 62.2 / 30.3 \\
FTK & Multiple-choice & Precision & 92.5 / \textbf{82.4} & 91.9 / 80.5 & 90.6 / 75.3 & \textbf{93.0} / 80.9 & 90.3 / 77.5 & 90.5 / 75.4 & 89.4 / 71.6 & 81.2 / 63.5 & 89.7 / 77.9 \\
FTK & Multiple-choice & Recall & 92.1 / 79.3 & 92.0 / 80.5 & 87.2 / 69.8 & 89.1 / 73.7 & 90.9 / 78.0 & 84.5 / 66.1 & 84.7 / 65.2 & 83.2 / 65.6 & \textbf{94.9} / \textbf{86.1} \\
FTK & Multiple-choice & F1 & 91.0 / 77.8 & 90.6 / 77.5 & 87.3 / 69.3 & 89.5 / 73.7 & 89.1 / 74.8 & 85.3 / 66.4 & 84.7 / 64.2 & 80.0 / 61.3 & \textbf{91.1} / \textbf{79.4} \\
FTK & Cloze & F1 & 58.2 / 20.9 & 58.8 / 19.7 & 54.9 / 19.8 & 59.9 / 23.5 & 56.1 / 17.5 & 50.2 / 16.2 & 56.0 / 15.9 & 42.1 / 14.0 & \textbf{60.2} / \textbf{23.9} \\
\midrule
CPMK & Cloze & F1 & 74.0 / \textbf{57.6} & \textbf{74.4} / 55.7 & 67.6 / 45.7 & 74.3 / 54.9 & 67.5 / 41.3 & 64.6 / 41.0 & 70.8 / 46.7 & 57.5 / 36.6 & 73.4 / 57.4 \\
CPMK & Single-choice & Accuracy & 74.3 / 31.0 & 73.3 / 25.5 & 67.3 / 18.0 & 68.6 / 17.8 & 66.9 / 16.3 & 60.5 / 12.8 & 63.1 / 8.0 & 44.5 / 13.0 & \textbf{78.1} / \textbf{38.3} \\
CPMK & Multiple-choice & Accuracy & \textbf{48.0} / \textbf{24.5} & 42.2 / 15.0 & 35.4 / 6.5 & 44.8 / 15.3 & 35.5 / 6.5 & 33.2 / 5.5 & 34.2 / 3.8 & 24.7 / 4.0 & 45.0 / 21.8 \\
CPMK & Multiple-choice & Precision & \textbf{86.4} / \textbf{73.5} & 84.3 / 69.6 & 82.4 / 62.7 & 86.1 / 69.7 & 81.9 / 62.1 & 83.1 / 63.5 & 80.6 / 58.7 & 76.0 / 53.9 & 84.1 / 71.8 \\
CPMK & Multiple-choice & Recall & 90.6 / 76.3 & 91.5 / 78.4 & 87.8 / 72.2 & 87.7 / 71.1 & 87.8 / 70.0 & 82.2 / 59.9 & 84.1 / 62.0 & 82.5 / 61.8 & \textbf{94.4} / \textbf{85.8} \\
CPMK & Multiple-choice & F1 & 87.0 / 72.4 & 86.3 / 71.5 & 83.3 / 64.6 & 85.4 / 68.0 & 83.2 / 63.8 & 80.6 / 58.6 & 80.1 / 57.0 & 77.3 / 55.6 & \textbf{87.7} / \textbf{76.6} \\
\midrule
IE & EMR & Precision & 66.6 / 46.7 & 64.6 / 46.7 & 72.8 / 56.2 & 69.1 / 50.5 & \textbf{73.3} / \textbf{61.3} & 61.8 / 45.5 & 38.1 / 17.3 & 59.7 / 41.4 & 72.6 / 58.3 \\
IE & EMR & Recall & 61.3 / 43.9 & 58.6 / 42.4 & \textbf{71.9} / \textbf{56.0} & 66.3 / 49.3 & 62.2 / 49.8 & 56.0 / 41.9 & 33.5 / 15.0 & 55.8 / 41.9 & 67.7 / 54.2 \\
IE & EMR & F1 & 63.1 / 44.2 & 60.8 / 43.4 & \textbf{71.7} / 55.1 & 67.2 / 49.1 & 66.7 / 54.2 & 58.1 / 42.6 & 35.2 / 15.5 & 57.1 / 40.6 & 69.6 / \textbf{55.5} \\
IE & Classical texts & Precision & 60.1 / 40.2 & \textbf{61.4} / \textbf{42.8} & 44.7 / 11.4 & 57.3 / 37.9 & 57.9 / 38.1 & 58.9 / 38.0 & 38.7 / 13.8 & 51.1 / 31.0 & 60.8 / 42.0 \\
IE & Classical texts & Recall & 62.9 / 44.7 & 64.4 / 48.8 & 51.7 / 16.3 & 62.6 / 45.9 & 56.9 / 37.9 & 61.6 / 42.6 & 40.3 / 14.9 & 56.6 / 38.5 & \textbf{65.9} / \textbf{51.8} \\
IE & Classical texts & F1 & 60.8 / 41.5 & 62.1 / 44.7 & 47.5 / 13.1 & 59.1 / 40.4 & 56.6 / 36.9 & 59.5 / 39.2 & 39.0 / 13.8 & 53.0 / 33.5 & \textbf{62.6} / \textbf{45.6} \\
\midrule
DTR & Syndrome & Precision & 9.9 / 1.3 & 11.0 / 1.5 & 5.0 / 0.7 & 9.0 / 1.2 & 7.7 / 0.6 & 9.1 / 1.2 & 10.7 / 1.5 & 3.2 / 0.6 & \textbf{12.3} / \textbf{2.5} \\
DTR & Syndrome & Recall & 13.3 / 1.7 & \textbf{14.3} / 1.8 & 8.4 / 1.1 & 13.4 / 1.9 & 11.2 / 0.9 & 12.6 / 1.6 & 12.6 / 1.7 & 4.5 / 0.7 & 13.0 / \textbf{2.5} \\
DTR & Syndrome & F1 & 10.9 / 1.5 & 11.9 / 1.6 & 6.0 / 0.9 & 10.4 / 1.4 & 8.8 / 0.7 & 10.1 / 1.4 & 11.0 / 1.6 & 3.6 / 0.6 & \textbf{12.1} / \textbf{2.4} \\
DTR & Treatment & Precision & 14.2 / 1.9 & 14.3 / 0.8 & 10.5 / 2.1 & 12.9 / 0.9 & 11.5 / 1.3 & 11.7 / 0.6 & 13.6 / 0.6 & 9.6 / 0.9 & \textbf{16.5} / \textbf{2.4} \\
DTR & Treatment & Recall & 17.0 / 2.3 & 16.4 / 0.8 & \textbf{19.2} / \textbf{4.6} & 15.6 / 1.2 & 14.4 / 1.5 & 15.0 / 0.9 & 14.5 / 0.9 & 11.2 / 1.2 & 15.6 / 2.0 \\
DTR & Treatment & F1 & 14.8 / 1.9 & 14.7 / 0.8 & 13.1 / \textbf{2.8} & 13.6 / 1.0 & 12.3 / 1.3 & 12.6 / 0.7 & 13.5 / 0.7 & 9.9 / 1.0 & \textbf{15.5} / 2.1 \\
DTR & Prescription & Precision & 35.7 / 19.7 & \textbf{36.9} / 19.9 & 27.6 / 17.1 & 33.4 / 19.0 & 31.0 / 17.0 & 32.1 / 18.0 & 32.4 / 17.4 & 30.2 / 17.3 & 36.6 / \textbf{24.3} \\
DTR & Prescription & Recall & 30.6 / 17.9 & 29.6 / 17.0 & \textbf{39.3} / 25.9 & 33.5 / 20.1 & 29.3 / 16.9 & 29.4 / 17.2 & 29.9 / 16.8 & 22.9 / 13.6 & 38.9 / \textbf{27.2} \\
DTR & Prescription & F1 & 32.3 / 18.2 & 32.1 / 17.7 & 31.9 / 20.1 & 32.7 / 18.9 & 29.5 / 16.4 & 30.0 / 17.0 & 30.4 / 16.4 & 25.4 / 14.7 & \textbf{37.0} / \textbf{24.9} \\
DTR & Dosage & MAE & 4.1 / 4.0 & 4.2 / 4.1 & 4.5 / 4.9 & 4.2 / 4.2 & 4.4 / 4.7 & 4.2 / 4.1 & \textbf{4.0} / 3.9 & 4.4 / \textbf{3.7} & 4.3 / 4.4 \\
DTR & Dosage & Cosine & 30.9 / 16.2 & 31.0 / 16.4 & 31.4 / 19.7 & 31.2 / 17.7 & 29.0 / 15.5 & 28.7 / 15.4 & 29.1 / 15.2 & 23.3 / 12.8 & \textbf{35.4} / \textbf{23.7} \\
\midrule
DTR-F1 & Syndrome & Precision & 20.4 / 8.9 & 21.5 / 10.0 & 12.1 / 5.8 & 19.0 / 9.5 & 16.5 / 7.8 & 18.1 / 7.9 & 20.9 / 9.3 & 10.6 / 5.4 & \textbf{25.1} / \textbf{12.8} \\
DTR-F1 & Syndrome & Recall & 26.6 / 11.2 & 27.1 / 11.2 & 20.1 / 9.1 & \textbf{27.4} / \textbf{12.3} & 23.8 / 10.5 & 24.8 / 10.3 & 24.1 / 9.7 & 14.9 / 6.8 & 25.5 / 12.1 \\
DTR-F1 & Syndrome & F1 & 22.0 / 9.6 & 22.9 / 10.1 & 14.6 / 6.9 & 21.6 / 10.4 & 18.7 / 8.7 & 20.0 / 8.5 & 21.3 / 9.1 & 11.9 / 5.8 & \textbf{24.2} / \textbf{11.9} \\
DTR-F1 & Treatment & Precision & 21.0 / 8.1 & 21.7 / 6.2 & 15.0 / 5.7 & 19.8 / 6.4 & 18.0 / 6.7 & 18.0 / 5.1 & 21.1 / 5.9 & 14.6 / 4.3 & \textbf{25.1} / \textbf{8.5} \\
DTR-F1 & Treatment & Recall & 25.2 / 10.6 & 24.9 / 7.8 & \textbf{28.1} / \textbf{12.3} & 24.1 / 8.7 & 23.0 / 9.0 & 23.3 / 7.3 & 22.5 / 7.2 & 17.7 / 5.9 & 23.7 / 8.4 \\
DTR-F1 & Treatment & F1 & 22.1 / \textbf{8.9} & 22.5 / 6.7 & 18.9 / 7.6 & 21.0 / 7.2 & 19.5 / 7.5 & 19.6 / 5.9 & 21.1 / 6.4 & 15.3 / 4.9 & \textbf{23.6} / 8.2 \\
DTR-F1 & Prescription & Precision & 35.6 / 19.8 & \textbf{37.0} / 20.0 & 27.7 / 17.3 & 33.4 / 19.2 & 30.9 / 17.0 & 31.6 / 18.0 & 33.5 / 17.9 & 30.1 / 17.4 & 36.7 / \textbf{24.3} \\
DTR-F1 & Prescription & Recall & 31.0 / 18.2 & 29.6 / 17.1 & \textbf{39.7} / 26.1 & 33.6 / 20.2 & 29.6 / 17.0 & 30.2 / 17.8 & 30.8 / 17.2 & 23.0 / 13.7 & 38.9 / \textbf{27.1} \\
DTR-F1 & Prescription & F1 & 32.4 / 18.4 & 32.2 / 17.8 & 32.0 / 20.3 & 32.7 / 19.0 & 29.6 / 16.5 & 30.2 / 17.3 & 31.2 / 16.9 & 25.5 / 14.8 & \textbf{37.1} / \textbf{25.0} \\
DTR-F1 & Dosage & MAE & 4.1 / 4.0 & 4.2 / 4.2 & 4.5 / 5.0 & 4.2 / 4.2 & 4.4 / 4.7 & 4.2 / 4.2 & \textbf{4.0} / 3.9 & 4.4 / \textbf{3.8} & 4.3 / 4.4 \\
DTR-F1 & Dosage & Cosine & 31.2 / 16.5 & 31.1 / 16.5 & 31.7 / 20.1 & 31.3 / 17.8 & 29.2 / 15.6 & 29.0 / 16.1 & 29.7 / 15.6 & 23.5 / 12.9 & \textbf{35.6} / \textbf{23.8} \\
\midrule
DR & Syndrome & Accuracy & 86.7 / 39.8 & 85.6 / 35.5 & 86.9 / 41.5 & 85.4 / 40.8 & 86.0 / 39.3 & 83.6 / 30.5 & 84.9 / 32.8 & 77.7 / 34.8 & \textbf{90.8} / \textbf{57.0} \\
DR & Treatment & Accuracy & 78.7 / 22.8 & 77.6 / 20.3 & 78.1 / 22.0 & 79.0 / 20.8 & 80.0 / 22.3 & 76.8 / 15.3 & 77.6 / 17.8 & 75.3 / 31.8 & \textbf{86.9} / \textbf{47.0} \\
DR & Prescription & Accuracy & 89.1 / 51.2 & 90.0 / 55.3 & 90.6 / 56.8 & 88.7 / 49.3 & 87.5 / 45.3 & 87.9 / 44.0 & 88.4 / 47.0 & 78.5 / 37.3 & \textbf{93.4} / \textbf{69.3} \\
\midrule
\multicolumn{2}{c}{\textbf{Overall Score}} &  & 64.6 / 38.9 & 64.4 / 38.2 & 61.9 / 34.1 & 64.1 / 37.5 & 62.3 / 34.5 & 60.1 / 30.9 & 58.7 / 28.7 & 52.1 / 26.9 & \textbf{67.1} / \textbf{45.3} \\
\bottomrule
\end{tabular}%
}
\end{table}

\begin{table}[!t]
\centering
\setlength{\tabcolsep}{2pt}
\renewcommand{\arraystretch}{0.68}
\caption{Primary LingLan results for TCM-oriented language models and related Chinese-language baselines. Cells report full-set / Hard-subset scores. The table includes the 16 metrics used to calculate Overall Score; bold marks the best value within each split.}
\label{tab:linglan_domain_matrix}
\resizebox{\linewidth}{!}{%
\begin{tabular}{lllcccccccc}
\toprule
\textbf{Dom.} & \textbf{Subtask} & \textbf{Metric} & \modelhead{Lingdan-V1}{13B} & \modelhead{TCMChat}{7B} & \modelhead{Baichuan2}{13B} & \modelhead{BianCang}{7B} & \modelhead{BianCang}{14B} & \modelhead{ShizhenGPT}{7B} & \modelhead{ShizhenGPT}{32B} & \modelhead{Lingdan-R1}{14B} \\
\midrule
TLE & Comprehensive & Accuracy & 44.3 / 20.8 & 31.6 / 19.5 & 50.1 / 21.8 & 88.9 / 70.0 & 91.2 / \textbf{73.5} & 83.6 / 54.5 & 90.7 / 67.0 & \textbf{91.7} / 67.0 \\
\midrule
FTK & Single-choice & Accuracy & 48.4 / 26.3 & 11.1 / 7.5 & 60.6 / 28.5 & 78.2 / 43.3 & 82.6 / 49.3 & 79.9 / 41.3 & 83.5 / 47.8 & \textbf{88.2} / \textbf{54.8} \\
FTK & Multiple-choice & F1 & 52.4 / 49.0 & 39.9 / 32.4 & 75.9 / 67.7 & 76.2 / 68.0 & 55.8 / 50.3 & 81.6 / 66.9 & 86.7 / 72.7 & \textbf{91.1} / \textbf{79.4} \\
FTK & Cloze & F1 & 8.5 / 4.5 & 13.9 / 6.5 & 33.2 / 13.3 & 35.4 / 15.0 & 38.7 / 14.6 & 40.6 / 17.7 & 49.8 / 20.9 & \textbf{60.2} / \textbf{23.9} \\
\midrule
CPMK & Single-choice & Accuracy & 28.3 / 16.3 & 19.2 / 18.5 & 41.6 / 24.8 & 58.4 / 31.8 & 62.5 / 35.5 & 56.5 / 20.3 & 68.7 / 27.3 & \textbf{78.1} / \textbf{38.3} \\
CPMK & Multiple-choice & F1 & 52.3 / 51.2 & 33.8 / 29.7 & 75.2 / 65.4 & 77.1 / 68.3 & 36.3 / 27.8 & 78.0 / 63.1 & 82.7 / 67.8 & \textbf{87.7} / \textbf{76.6} \\
CPMK & Cloze & F1 & 11.7 / 9.3 & 13.8 / 10.6 & 37.2 / 26.4 & 54.4 / 37.5 & 58.4 / 42.2 & 55.2 / 42.5 & 69.3 / 53.6 & \textbf{73.4} / \textbf{57.4} \\
\midrule
IE & EMR & F1 & 0.8 / 0.8 & 0.0 / 0.0 & 25.1 / 22.7 & 16.8 / 12.9 & 29.5 / 24.8 & 29.1 / 27.4 & 52.4 / 46.4 & \textbf{69.6} / \textbf{55.5} \\
IE & Classical texts & F1 & 3.4 / 1.7 & 0.3 / 0.4 & 29.0 / 18.7 & 44.1 / 24.5 & 49.0 / 34.0 & 38.1 / 22.1 & 53.0 / 34.5 & \textbf{62.6} / \textbf{45.6} \\
\midrule
DTR & Syndrome & F1 & 0.7 / 0.0 & 0.7 / 0.3 & 4.3 / 1.3 & 4.9 / 1.1 & 7.3 / 1.2 & 8.6 / 1.8 & 9.0 / 2.0 & \textbf{12.1} / \textbf{2.4} \\
DTR & Treatment & F1 & 1.0 / 0.1 & 2.8 / 0.6 & 7.4 / 1.3 & 7.1 / 1.1 & 9.1 / 1.2 & 11.9 / 2.0 & 12.3 / \textbf{2.1} & \textbf{15.5} / \textbf{2.1} \\
DTR & Prescription & F1 & 0.2 / 0.2 & 0.0 / 0.0 & 13.6 / 8.3 & 14.4 / 9.1 & 8.2 / 5.1 & 25.5 / 16.0 & 27.0 / 16.8 & \textbf{37.0} / \textbf{24.9} \\
DTR & Dosage & Cosine & 0.2 / 0.1 & 0.0 / 0.0 & 12.9 / 7.5 & 13.5 / 8.6 & 7.5 / 4.9 & 25.2 / 15.2 & 26.7 / 15.9 & \textbf{35.4} / \textbf{23.7} \\
\midrule
DR & Syndrome & Accuracy & 32.5 / 17.3 & 47.2 / 28.5 & 64.6 / 32.0 & 73.8 / 34.3 & 80.7 / 41.5 & 78.8 / 35.8 & 82.0 / 33.5 & \textbf{90.8} / \textbf{57.0} \\
DR & Treatment & Accuracy & 38.5 / 20.5 & 48.3 / 27.5 & 62.6 / 28.5 & 72.0 / 32.3 & 72.7 / 26.5 & 72.4 / 21.8 & 74.9 / 21.0 & \textbf{86.9} / \textbf{47.0} \\
DR & Prescription & Accuracy & 18.1 / 15.8 & 23.0 / 18.5 & 36.2 / 25.3 & 65.6 / 39.8 & 80.5 / 48.0 & 77.9 / 40.0 & 87.9 / 52.0 & \textbf{93.4} / \textbf{69.3} \\
\midrule
\multicolumn{2}{c}{\textbf{Overall Score}}  &  & 21.3 / 14.6 & 17.9 / 12.5 & 39.3 / 24.6 & 48.8 / 31.1 & 48.1 / 30.0 & 52.7 / 30.5 & 59.8 / 36.3 & \textbf{67.1} / \textbf{45.3} \\
\bottomrule
\end{tabular}%
}
\end{table}

\subsection{GRPO-associated Changes Across Model Scales}

\begin{figure}[!htbp]
    \centering
    \includegraphics[width=\linewidth]{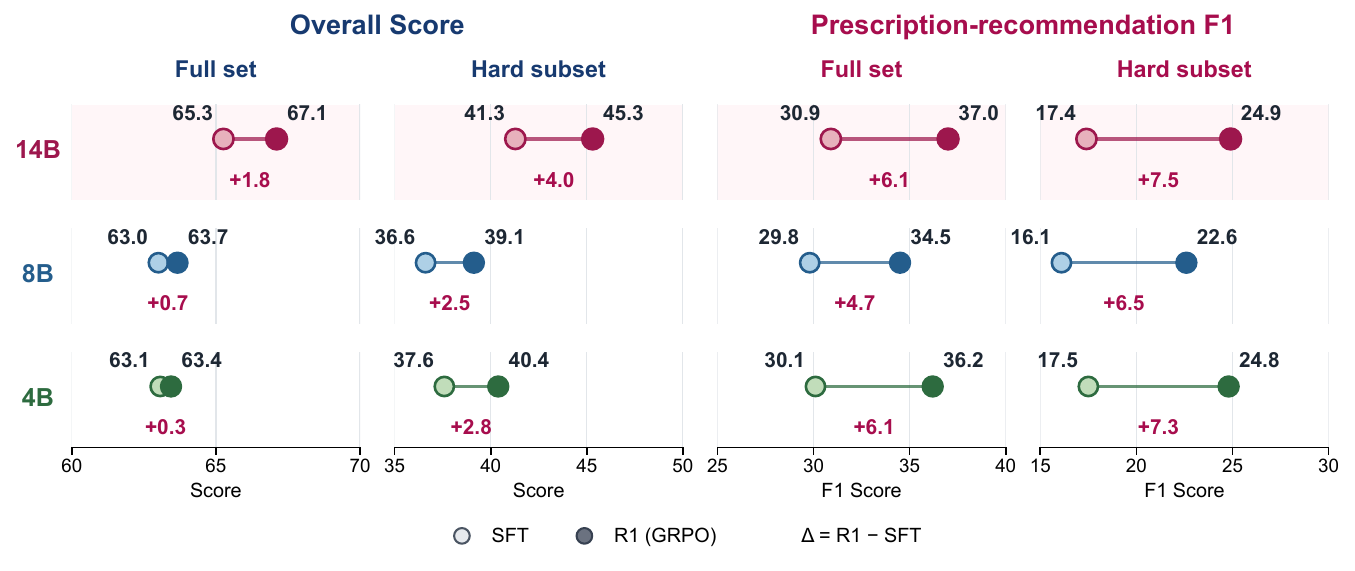}
    \caption{Performance changes from the SFT to R1 checkpoints across Lingdan-V2 model scales. Each cell compares the corresponding SFT and R1 checkpoints for one evaluation metric and benchmark split; endpoint labels show the two scores, and $\Delta$ denotes R1 minus SFT.}
    \label{fig:lingdan_training_stage}
\end{figure}

We compared each Lingdan-SFT checkpoint with its corresponding Lingdan-R1 checkpoint to characterize the performance changes associated with prescription-centered GRPO (Figure~\ref{fig:lingdan_training_stage}). In the primary 14B lineage, Overall Score increased from 65.3 to 67.1 on the full LingLan benchmark and from 41.3 to 45.3 on LingLan-Hard. Prescription-recommendation F1 increased from 30.9 to 37.0 and from 17.4 to 24.9 on the two evaluation splits, respectively. For both metrics, the absolute improvement was larger on LingLan-Hard than on the full benchmark.

The improvement in prescription recommendation was consistent across model scales. On the full benchmark, prescription-recommendation F1 increased from 30.1 to 36.2 at 4B, from 29.8 to 34.5 at 8B, and from 30.9 to 37.0 at 14B; the corresponding changes on LingLan-Hard were 17.5 to 24.8, 16.1 to 22.6, and 17.4 to 24.9. Overall Score also increased at every model scale on both evaluation splits. At 4B, 8B, and 14B, the respective changes were 63.1 to 63.4, 63.0 to 63.7, and 65.3 to 67.1 on the full benchmark, and 37.6 to 40.4, 36.6 to 39.1, and 41.3 to 45.3 on LingLan-Hard. The larger gains in prescription-recommendation F1 indicate that the most pronounced cross-scale change remained concentrated in the task directly represented by the GRPO reward.

\subsection{Herbal Prescription Recommendation}

We next evaluated prescription recommendation on the private spleen-stomach disease evaluation set to examine whether the performance observed on LingLan was also evident in retrospective clinical cases (Table~\ref{tab:pwb_results}; Figure~\ref{fig:pwb_prescription}). Lingdan-14B-R1 achieved a herb-level F1 of 52.8, a dose cosine similarity of 50.6, and a dose MAE of 4.369. In comparison, Lingdan-14B-SFT obtained 35.1, 34.8, and 5.840 on the three metrics, respectively. The increase in herb-level F1 was primarily associated with higher recall, which increased from 27.1 to 55.4, while precision remained similar (52.0 versus 51.8). Lingdan-14B-R1 also outperformed Qwen3-14B, which achieved an F1 of 33.3, a dose cosine similarity of 33.6, and a dose MAE of 5.737, as well as Qwen3-32B, which achieved 32.9, 33.2, and 5.812, respectively.

Higher herb-composition agreement after GRPO was also observed at the smaller model scales. At 4B, herb-level F1 increased from 33.7 for Lingdan-4B-SFT to 52.7 for Lingdan-4B-R1, while dose cosine similarity increased from 34.1 to 49.5. At 8B, the corresponding metrics increased from 34.2 to 50.7 and from 34.2 to 49.5. Changes in absolute dosage error were less consistent: dose MAE decreased from 5.787 to 5.062 at 4B but increased from 5.812 to 6.006 at 8B. Thus, the improvement in herb composition was reproduced across model scales, whereas dosage estimation did not improve uniformly.
Overall, Lingdan-14B-R1 achieved the highest herb-level F1 and dose cosine similarity and the lowest dose MAE among the evaluated models. These results show stronger agreement with reference prescriptions under the defined retrospective, text-based evaluation protocol.

\begin{table}[!htbp]
    \centering
    \caption{Prescription recommendation performance on the private spleen-stomach disease test set. Herb-level metrics and dose cosine similarity are reported on a 0--100 scale; lower dose MAE indicates better dosage estimation.}
    \label{tab:pwb_results}
    \begin{tabular}{lccccc}
    \toprule
    Model & F1 & Precision & Recall & Dose cosine & Dose MAE \\
    \midrule
    Lingdan-14B-R1  & \textbf{52.8} & 51.8 & \textbf{55.4} & \textbf{50.6} & \textbf{4.369} \\
    Lingdan-8B-R1   & 50.7 & \textbf{55.0} & 48.3 & 49.5 & 6.006 \\
    Lingdan-4B-R1   & 52.7 & 52.5 & 54.1 & 49.5 & 5.062 \\
    Lingdan-14B-SFT & 35.1 & 52.0 & 27.1 & 34.8 & 5.840 \\
    Lingdan-8B-SFT  & 34.2 & 52.7 & 26.0 & 34.2 & 5.812 \\
    Lingdan-4B-SFT  & 33.7 & 51.6 & 25.7 & 34.1 & 5.787 \\
    Qwen3-4B        & 30.7 & 49.2 & 22.9 & 32.4 & 5.382 \\
    Qwen3-8B        & 34.2 & 51.5 & 26.2 & 34.6 & 6.126 \\
    Qwen3-14B       & 33.3 & 49.0 & 25.8 & 33.6 & 5.737 \\
    Qwen3-32B       & 32.9 & 49.2 & 25.3 & 33.2 & 5.812 \\
    ShizhenGPT-7B   & 30.8 & 49.8 & 22.8 & 33.6 & 6.884 \\
    ShizhenGPT-32B  & 30.0 & 46.0 & 22.7 & 31.3 & 6.114 \\
    \bottomrule
    \end{tabular}
\end{table}

\subsection{Comparison with Existing TCM-oriented Language Models}

We further positioned Lingdan-14B-R1 against existing TCM-oriented language models and a general Chinese-language baseline (Table~\ref{tab:linglan_domain_matrix}). Lingdan-14B-R1 achieved Overall Scores of 67.1 on the full LingLan benchmark and 45.3 on LingLan-Hard, exceeding the strongest evaluated domain-specific baseline, ShizhenGPT-32B, by 7.3 and 9.0 points, respectively. Among the models included in this comparison, Lingdan-14B-R1 obtained the highest score on all 16 primary metrics on the full benchmark and the highest or tied-highest score on 15 of the 16 metrics on LingLan-Hard. The main exception was TLE accuracy on LingLan-Hard, where BianCang-14B achieved 73.5 compared with 67.0 for Lingdan-14B-R1.

The advantage was also evident on the private spleen-stomach disease prescription evaluation (Table~\ref{tab:pwb_results}). Lingdan-14B-R1 ranked first in herb-level F1 and dose cosine similarity, with scores of 52.8 and 50.6, respectively, while achieving the lowest dose MAE of 4.369. Among the non-Lingdan models, the strongest results were 34.2 for herb-level F1, 34.6 for dose cosine similarity, and 5.382 for dose MAE. These comparisons place Lingdan-14B-R1 ahead of the evaluated TCM-oriented and general Chinese-language baselines across both the comprehensive LingLan benchmark and the separate prescription-recommendation evaluation.

\section{Discussion}


Lingdan-V2 advances TCM language modeling through the coordinated scaling of domain-specific data, reasoning-oriented supervision, and prescription-centered reinforcement post-training. By integrating large-scale TCM clinical corpora, complementary reasoning supervision, and prescription-oriented optimization, Lingdan-V2 establishes a multi-stage framework for developing TCM language models with broad domain knowledge and specialized clinical generation capabilities. The results demonstrate that domain adaptation and reasoning supervision provide the foundation for general TCM capabilities, while prescription-centered GRPO further improves prescription-related performance across model scales, benchmark difficulties, and retrospective clinical evaluations.

\subsection{Domain Grounding and Reasoning Capability Formation}

Lingdan-V2 was designed around the premise that effective TCM language models require both domain grounding and structured reasoning capability. The continued-pretraining corpus contained 16.62 billion TCM-related tokens, including approximately 10.4 billion tokens from clinical corpora. The large proportion of clinical text was intended to expose the model to the distribution of real-world TCM practice, where symptom descriptions, syndrome formulations, treatment principles, and prescription records are expressed with substantial variability. Such clinical exposure provides a foundation for connecting formal TCM knowledge with the language patterns required for downstream diagnostic and prescription-oriented tasks.
Beyond domain knowledge acquisition, Lingdan-V2 incorporated reasoning-oriented supervision to bridge general language modeling and clinical reasoning. The SFT stage combined TCM-specific reasoning distillation with general reasoning supervision: approximately 240K TCM reasoning samples distilled from DeepSeek-R1 \cite{guo2025deepseek} were used to strengthen structured clinical reasoning, while 110K general reasoning samples from Qwen3-235B \cite{yang2025qwen3} were included to preserve broader reasoning behaviors inherited from the Qwen3 backbone. This combination reflects a balance between domain specialization and capability retention, allowing the model to acquire TCM-specific reasoning patterns without relying exclusively on a narrow clinical distribution.
Lingdan-V2 was trained using full-parameter optimization throughout the training pipeline. This design enables domain adaptation and task supervision to update the entire model representation, although it also introduces substantially higher computational costs than parameter-efficient alternatives.

\subsection{Prescription-centered GRPO and Task-specific Alignment}

Prescription-centered GRPO was introduced to bridge the gap between acquired TCM knowledge and clinical prescription decision making. TCM prescriptions represent structured clinical decisions that integrate syndrome differentiation, treatment principles, herb selection, and dosage patterns. By optimizing against prescription-level feedback, GRPO encouraged the model to transform previously acquired domain knowledge and reasoning capability into more consistent prescription-oriented outputs. The SFT-to-R1 comparisons demonstrate that prescription-centered GRPO produced its most consistent improvements on prescription-related tasks. Prescription-recommendation performance increased across the 4B, 8B, and 14B model scales on both LingLan and LingLan-Hard (Figure~\ref{fig:lingdan_training_stage}).
In the 14B lineage, prescription generation showed the largest improvement, accompanied by gains in Overall Score on both evaluation splits. The larger improvement on LingLan-Hard suggests that prescription-oriented reinforcement optimization may be particularly beneficial under more challenging evaluation settings.
The effects of GRPO were not uniformly distributed across all capabilities. While Overall Score and prescription recommendation improved consistently across model scales, changes in individual knowledge, extraction, reasoning, and dosage metrics varied across tasks. In the primary 14B lineage, most evaluated capabilities improved after GRPO, whereas clinical-record information extraction showed a slight decrease. This pattern may reflect a shift toward structured clinical reasoning and longer-form generation, which can introduce a trade-off with tasks requiring concise and precise information extraction. These findings indicate that reinforcement post-training can effectively optimize targeted clinical behaviors, while its broader effects depend on the interaction between reward design, model initialization, and evaluation objectives.

\subsection{Physician-source Heterogeneity and Reward Stability}

\begin{figure}[!htbp]
    \centering
    \includegraphics[width=\linewidth]{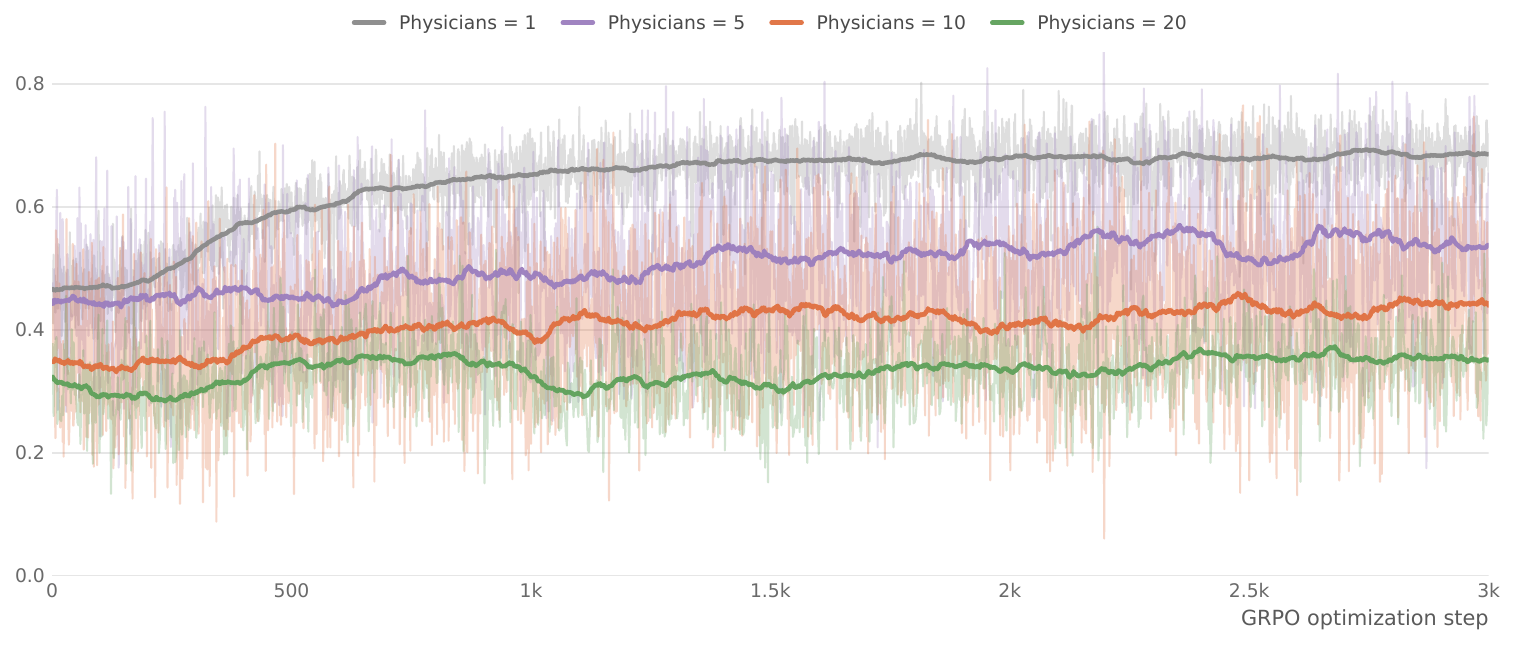}
    \caption{GRPO reward trajectories under different physician-source configurations. The curves compare prescription training pools containing cases from 1, 5, 10, and 20 physicians over the first 3,000 global optimization steps. Faint curves show raw rewards, and solid curves show the corresponding moving averages.}
    \label{fig:physician_source_dynamics}
\end{figure}

Physician-source heterogeneity had a pronounced influence on GRPO optimization. We compared prescription training pools containing cases from 1, 5, 10, and 20 physicians. As shown in Figure~\ref{fig:physician_source_dynamics}, the single-physician configuration achieved the strongest reward improvement, followed by the five-physician configuration, whereas the 10- and 20-physician configurations exhibited progressively weaker optimization trends. Additional experiments under the 10-physician configuration produced similar patterns across different training settings (Figure~\ref{fig:appendix_top10_runs}), indicating that the observed optimization difficulty could not be readily resolved through hyperparameter adjustment.
This result highlights a fundamental tension between expert diversity and reward consistency in prescription-centered reinforcement learning. TCM prescriptions are highly individualized clinical decisions. Even for similar clinical presentations, physicians may differ in syndrome interpretation, treatment strategy, formula modification, herb selection, and dosage practice. Consequently, broader physician-source coverage increases the diversity of clinically reasonable prescription targets, while making it more difficult for a reference-based reward function to provide a consistent optimization signal.
The final GRPO corpus comprised six physician sources. It combined 9,849 augmented physician-style cases derived from five physicians with 14,363 cleaned prescription cases contributed by an additional spleen-stomach disease specialist.

\subsection{Clinical Interpretation of Prescription Composition and Dosage}

The private spleen-stomach disease evaluation provides a clinically grounded perspective on how prescription-centered optimization affects structured prescription generation. Beyond herb-level agreement, dosage evaluation reveals complementary aspects of prescription quality. Dose cosine similarity reflects the relative structure of the predicted dosage distribution, whereas dose MAE measures absolute deviation from reference amounts. The improvement observed across these metrics suggests that Lingdan-14B-R1 achieved closer agreement with reference dosage patterns from both relative and absolute perspectives. Together with herb-level matching, these complementary signals provide a more comprehensive and interpretable evaluation framework for prescription generation in clinical language models.

\subsection{Evolution from Lingdan-V1 and Positioning among Existing TCM Language Models}

Lingdan-V1 established the feasibility of adapting a general language model to TCM diagnosis, treatment, and prescription tasks through continued pretraining and supervised fine-tuning \cite{hua2024lingdan}. Lingdan-V2 substantially extends this foundation through a major expansion of training scale and clinical coverage. The training data used for Lingdan-V1 represented approximately 1.9\% of the total volume used for Lingdan-V2. This expansion enabled broader exposure to TCM knowledge and clinical language, strengthening the connection between formal domain knowledge and the diagnostic and prescription patterns encountered in clinical practice.
The evolution from Lingdan-V1 to Lingdan-V2 extends beyond data scaling. Lingdan-V2 integrates domain grounding, reasoning-oriented supervision, and prescription-centered reinforcement post-training within a model family spanning three parameter scales. Its evaluation framework further examines changes between the SFT and R1 stages, performance under increased benchmark difficulty, the influence of physician-source heterogeneity on reward stability, and prescription recommendation in retrospective clinical cases. These developments extend the Lingdan framework from domain adaptation toward the systematic optimization and evaluation of clinically oriented TCM reasoning and generation.

Among the evaluated TCM-oriented language models, Lingdan-14B-R1 demonstrated broader and more balanced performance across TCM knowledge, information extraction, diagnostic and therapeutic reasoning, decision recognition, and prescription recommendation (Table~\ref{tab:linglan_domain_matrix}). Although some existing models retained advantages on individual tasks, Lingdan-14B-R1 showed greater consistency across capability domains and maintained its advantage on the private spleen-stomach disease evaluation.

\subsection{Limitations and Future Work}

The current study focuses on text-based models with parameter scales up to 14B. 
Although clinical text captures substantial diagnostic and therapeutic information, it cannot directly represent tongue images, pulse signals, medical imaging, or other multimodal evidence used in clinical decision making. 
Future versions of Lingdan will integrate heterogeneous clinical modalities and explore larger dense or mixture-of-experts architectures to further expand model capacity and support more comprehensive TCM reasoning.
A further direction is to move beyond static input-output modeling toward clinical world models that represent disease progression, treatment interventions, and longitudinal outcomes. 
Such models could support reasoning over evolving patient states, simulate the potential consequences of alternative treatment strategies, and improve long-horizon clinical planning. 

\section{Conclusion}

Lingdan-V2 establishes a multi-scale TCM language model family through large-scale domain adaptation, reasoning-oriented supervision, and prescription-centered reinforcement post-training. Lingdan-14B-R1 achieved strong and balanced performance across the full LingLan benchmark and LingLan-Hard, while showing particularly clear improvements in prescription generation and retrospective clinical prescription recommendation.
The study also identified physician-source heterogeneity as an important factor affecting reward stability in clinical reinforcement learning. Together, these findings demonstrate the value of combining large-scale TCM clinical data, structured reasoning supervision, and prescription-oriented optimization for developing clinically capable TCM language models. Future development will extend Lingdan toward larger model architectures, multimodal clinical inputs, and world-model approaches capable of representing longitudinal disease progression, treatment interventions, and clinical outcomes.

\section*{Data availability}
Publicly released data and evaluation resources associated with this study are available through the Lingdan-V2 project repository at \url{https://github.com/TCMAI-BJTU/Lingdan-V2}. The private clinical data cannot be publicly released because they contain sensitive clinical information.

\section*{Code and model availability}
The Lingdan-V2 models and associated resources are available through the project repository at \url{https://github.com/TCMAI-BJTU/Lingdan-V2}.

\section*{CRediT authorship contribution statement}

\textbf{Rui Hua}: Conceptualization, Methodology, Investigation, Formal analysis, Visualization, Writing -- original draft.
\textbf{Yu Wei}: Data curation, Resources.
\textbf{Ziri Jin}: Data curation, Investigation.
\textbf{Zhuo Sun}: Data curation, Investigation.
\textbf{Kai Chang}: Software, Investigation, Validation.
\textbf{Jianan Xia}: Writing -- review \& editing, Validation.
\textbf{Zixin Shu}: Data curation, Resources.
\textbf{Xiaodong Li}: Validation, Investigation.
\textbf{Dongmei Jia}: Data curation, Validation, Resources.
\textbf{Fei Dong}: Data curation, Validation, Resources.
\textbf{Runshun Zhang}: Data curation, Validation, Resources.
\textbf{Jian Yu}: Methodology, Formal analysis.
\textbf{Hao Xu}: Data curation, Validation, Resources.
\textbf{Haibin Yu}: Data curation, Validation, Resources.
\textbf{Jiansheng Li}: Data curation, Validation, Supervision.
\textbf{Baoyan Liu}: Supervision, Writing -- review \& editing.
\textbf{Wenjia Wang}: Supervision, Project administration, Writing -- review \& editing.
\textbf{Xuezhong Zhou}: Conceptualization, Supervision, Project administration, Funding acquisition, Writing -- review \& editing.

\section*{Funding}
This work is partially supported by the National Key Research and Development Program (No. 2023YFC3502604), the Natural Science Foundation of Beijing (No. L232033), the National Natural Science Foundation of China (Nos. 82374302, 82274352, 82505376 and U23B2062), the Fundamental Research Funds for the Central Universities (No. 2025JBZX064).

The funders had no role in the study design, data collection,
data analysis or interpretation, manuscript preparation, or
the decision to submit the article for publication.

\section*{Declaration of competing interest}
We declare that none of the authors have competing financial or nonfinancial interests.

\bibliographystyle{cas-model2-names}
\bibliography{cas-refs}
\clearpage
\begin{appendices}

\numberwithin{table}{section}
\numberwithin{figure}{section}
\numberwithin{equation}{section}

\section{Public Training Data Sources}

\begin{table}[pos=!hbp]
    \centering
    \caption{Public datasets used for Lingdan-V2 training.}
    \label{tab:public_datasets}
    \begin{threeparttable}
    \begin{tabular}{ll}
    \toprule
    Dataset & Category \\
    \midrule
    TCMChat \cite{dai2024tcmchat} & TCM dialogue \\
    Huatuo-26M \cite{li2023huatuo26m} & Medical instruction data \\
    CMB \cite{wang2024cmb} & Chinese medical knowledge benchmark \\
    CMExam \cite{liu2023benchmarking} & TCM examination dataset \\
    MTCMB \cite{kong2025mtcmb} & TCM multi-task dataset \\
    cMedQA2 \cite{zhang2018cmedqa2} & Medical question answering \\
    MLEC-QA \cite{li2021mlecqa} & Medical examination dataset \\
    TCM-SD \cite{ren2022tcmsd} & TCM syndrome differentiation dataset \\
    RJUA-QA \cite{lyu2023rjuaqa} & Medical question answering \\
    MedDialog \cite{zeng2020meddialog} & Medical dialogue dataset \\
    Traditional Chinese Medicine Dataset Pretrain\tnote{a} & TCM pretraining corpus \\
    Traditional Chinese Medicine Dataset SFT\tnote{b} & TCM instruction data \\
    Traditional Chinese Medicine Exam\tnote{c} & TCM examination dataset \\
    ShenNong TCM Dataset\tnote{d} & TCM instruction data \\
    Chinese Medical Dialogue Data\tnote{e} & Medical dialogue dataset \\
    FineTome-100k\tnote{f} & General instruction following \\
    OpenMathReasoning \cite{moshkov2025openmathreasoning} & Mathematical reasoning \\
    DeepSeek-R1 distilled data \cite{guo2025deepseek}\tnote{g} & Chinese chain-of-thought reasoning \\
    Chinese Qwen3 reasoning data\tnote{h} & Chinese chain-of-thought reasoning \\
    \bottomrule
    \end{tabular}
    \begin{tablenotes}[flushleft]
    \footnotesize
    \item[a] \url{https://huggingface.co/datasets/SylvanL/Traditional-Chinese-Medicine-Dataset-Pretrain}
    \item[b] \url{https://huggingface.co/datasets/SylvanL/Traditional-Chinese-Medicine-Dataset-SFT}
    \item[c] \url{https://huggingface.co/datasets/SylvanL/Traditional-Chinese-Medicine-Exam}
    \item[d] \url{https://huggingface.co/datasets/michaelwzhu/ShenNong_TCM_Dataset}
    \item[e] \url{https://github.com/Toyhom/Chinese-medical-dialogue-data}
    \item[f] \url{https://huggingface.co/datasets/mlabonne/FineTome-100k}
    \item[g] \url{https://huggingface.co/datasets/Congliu/Chinese-DeepSeek-R1-Distill-data-110k}
    \item[h] \url{https://modelscope.cn/datasets/swift/Chinese-Qwen3-235B-Thinking-2507-Distill-data-110k-SFT}
    \end{tablenotes}
    \end{threeparttable}
\end{table}

\clearpage
\section{GRPO Optimization Details}
\label{app:grpo_details}

\subsection{Group-relative Policy Optimization}

For a prompt $x$, let $y_i$ denote the $i$th completion in a group of $G=8$ completions and let $r_i$ denote its composite reward. The group-relative advantage was defined as

\begin{equation}
\hat{A}_i = \frac{r_i-\bar{r}}{s_r},
\end{equation}

where $\bar{r}$ and $s_r$ denote the mean and standard deviation of rewards within the sampled completion group. Groups with zero reward variance were dynamically resampled up to five times.

With $\beta=0$, no KL regularization was applied. The clipped GRPO objective was defined as

\begin{equation}
    \mathcal{L}_{\mathrm{GRPO}}(\theta)
    =
    -\mathbb{E}
    \left[
    \frac{1}{G}
    \sum_{i=1}^{G}
    \sum_t
    m_{i,t}
    \min
    \left(
    \rho_{i,t}(\theta)\hat{A}_i,
    \mathrm{clip}
    \left(
    \rho_{i,t}(\theta),
    1-\epsilon_{\mathrm{low}},
    1+\epsilon_{\mathrm{high}}
    \right)
    \hat{A}_i
    \right)
    \right],
\end{equation}

where

\[
\rho_{i,t}(\theta)=
\frac{\pi_{\theta}(y_{i,t}\mid x,y_{i,<t})}
{\pi_{\theta_{\mathrm{old}}}(y_{i,t}\mid x,y_{i,<t})},
\]

$m_{i,t}$ denotes the sequence mask for valid completion tokens, and $\epsilon_{\mathrm{low}}=0.20$ and $\epsilon_{\mathrm{high}}=0.25$.
\subsection{Length Penalty}
For a completion of length $l$, the soft length penalty was defined as

\begin{equation}
r_{\mathrm{length}}(l)=
\min\left(
-\frac{l-(L_{\max}-L_{\mathrm{cache}})}
{L_{\mathrm{cache}}},
0
\right),
\end{equation}

where $L_{\max}=3{,}072$ and $L_{\mathrm{cache}}=128$. The length penalty remained zero up to 2,944 completion tokens and became increasingly negative thereafter.

\clearpage
\section{Complete LingLan Results for Smaller Models}

\begin{table}[pos=!hbp]
    \centering
    \setlength{\tabcolsep}{2pt}
    \renewcommand{\arraystretch}{0.68}
    \caption{Complete LingLan task-level results for smaller external models and the Lingdan-V2 R1 family at 4B, 8B and 14B. Qwen3-32B is repeated as a scale-boundary anchor. Cells report full-set / Hard-subset scores; bold marks the best value within this table and split, with lower values better for MAE. The final row reports Overall Score, as defined in Methods.}
    \label{tab:linglan_small_matrix}
    \resizebox{\linewidth}{!}{%
    \begin{tabular}{lllcccccccccc}
    \toprule
    \textbf{Dom.} & \textbf{Subtask} & \textbf{Metric} & \modelhead{Qwen3}{32B} & \modelhead{Qwen3}{30B-A3B} & \modelhead{Qwen3}{14B} & \modelhead{Qwen3}{8B} & \modelhead{Qwen3}{4B} & \modelhead{GPT-5-mini}{Undisclosed} & \modelhead{GPT-OSS}{20B} & \modelhead{Lingdan-4B}{R1} & \modelhead{Lingdan-8B}{R1} & \modelhead{Lingdan-14B}{R1} \\
    \midrule
    TLE & Comprehensive & Accuracy & 91.2 / 62.3 & 88.8 / 55.3 & 87.6 / 50.0 & 84.6 / 40.5 & 76.6 / 30.3 & 76.3 / 24.5 & 48.3 / 13.5 & 81.0 / 40.8 & 86.2 / 50.2 & \textbf{91.7} / \textbf{67.0} \\
    \midrule
    FTK & Single-choice & Accuracy & 86.1 / 44.3 & 82.6 / 32.0 & 83.0 / 34.5 & 80.8 / 33.8 & 76.2 / 28.5 & 77.0 / 28.7 & 65.4 / 29.3 & 81.5 / 40.3 & 83.9 / 46.0 & \textbf{88.2} / \textbf{54.8} \\
    FTK & Multiple-choice & Accuracy & 56.3 / 23.3 & 52.0 / 17.0 & 49.5 / 13.8 & 47.8 / 13.0 & 42.4 / 12.3 & 46.2 / 10.8 & 24.8 / 6.0 & 51.6 / 23.0 & 55.6 / 24.5 & \textbf{62.2} / \textbf{30.3} \\
    FTK & Multiple-choice & Precision & 90.3 / 77.5 & 89.8 / 75.2 & \textbf{90.6} / 75.2 & 88.4 / 71.8 & 86.1 / 69.0 & 86.1 / 67.7 & 76.1 / 59.5 & 86.2 / 73.2 & 90.1 / 77.5 & 89.7 / \textbf{77.9} \\
    FTK & Multiple-choice & Recall & 90.9 / 78.0 & 88.3 / 73.5 & 85.7 / 69.4 & 86.8 / 70.2 & 84.9 / 67.6 & 86.8 / 69.0 & 74.9 / 58.1 & 93.6 / 85.3 & 89.5 / 78.8 & \textbf{94.9} / \textbf{86.1} \\
    FTK & Multiple-choice & F1 & 89.1 / 74.8 & 87.4 / 71.0 & 86.3 / 68.7 & 85.8 / 67.6 & 83.5 / 64.8 & 84.8 / 65.3 & 72.9 / 55.0 & 88.4 / 76.3 & 88.2 / 74.9 & \textbf{91.1} / \textbf{79.4} \\
    FTK & Cloze & F1 & 56.1 / 17.5 & 53.9 / 17.9 & 54.1 / 14.3 & 51.1 / 12.3 & 49.0 / 15.9 & 49.2 / 17.9 & 35.9 / 10.6 & 53.4 / 19.3 & 55.2 / 17.4 & \textbf{60.2} / \textbf{23.9} \\
    \midrule
    CPMK & Cloze & F1 & 67.5 / 41.3 & 66.3 / 41.3 & 66.8 / 39.8 & 64.1 / 37.9 & 61.4 / 37.0 & 65.1 / 43.3 & 49.9 / 29.5 & 67.3 / 45.6 & 67.6 / 45.0 & \textbf{73.4} / \textbf{57.4} \\
    CPMK & Single-choice & Accuracy & 66.9 / 16.3 & 58.4 / 5.5 & 58.9 / 9.8 & 56.6 / 9.0 & 51.3 / 9.3 & 51.8 / 9.5 & 37.6 / 11.8 & 66.3 / 24.5 & 69.8 / 24.3 & \textbf{78.1} / \textbf{38.3} \\
    CPMK & Multiple-choice & Accuracy & 35.5 / 6.5 & 34.7 / 5.8 & 34.4 / 5.3 & 31.4 / 7.2 & 28.7 / 4.5 & 28.9 / 4.0 & 18.5 / 3.8 & 33.1 / 12.0 & 38.6 / 13.8 & \textbf{45.0} / \textbf{21.8} \\
    CPMK & Multiple-choice & Precision & 81.9 / 62.1 & 81.8 / 60.6 & 82.5 / 61.5 & 81.3 / 61.6 & 79.3 / 59.3 & 78.2 / 54.9 & 72.6 / 51.1 & 80.0 / 64.4 & 83.1 / 66.3 & \textbf{84.1} / \textbf{71.8} \\
    CPMK & Multiple-choice & Recall & 87.8 / 70.0 & 85.4 / 65.1 & 84.9 / 64.6 & 84.2 / 66.3 & 84.7 / 67.7 & 82.7 / 57.9 & 75.4 / 55.4 & 90.6 / 78.4 & 87.1 / 73.2 & \textbf{94.4} / \textbf{85.8} \\
    CPMK & Multiple-choice & F1 & 83.2 / 63.8 & 82.0 / 60.7 & 82.0 / 60.4 & 80.9 / 61.6 & 80.0 / 60.7 & 78.7 / 54.3 & 71.6 / 50.8 & 83.4 / 68.8 & 83.6 / 67.6 & \textbf{87.7} / \textbf{76.6} \\
    \midrule
    IE & EMR & Precision & 73.3 / 61.3 & 69.3 / 57.2 & 74.6 / 62.8 & 67.7 / 55.1 & 64.7 / 52.7 & 52.1 / 35.2 & 55.3 / 39.5 & 74.0 / 62.5 & \textbf{77.2} / \textbf{66.7} & 72.6 / 58.3 \\
    IE & EMR & Recall & 62.2 / 49.8 & 48.4 / 37.9 & 62.1 / 52.0 & 57.6 / 45.1 & 52.3 / 41.3 & 57.0 / 43.6 & 51.6 / 38.5 & 68.1 / 56.0 & \textbf{74.1} / \textbf{63.9} & 67.7 / 54.2 \\
    IE & EMR & F1 & 66.7 / 54.2 & 56.2 / 44.6 & 67.1 / 55.7 & 61.5 / 48.6 & 57.2 / 45.4 & 54.0 / 38.3 & 52.8 / 38.0 & 70.5 / 58.4 & \textbf{75.3} / \textbf{64.6} & 69.6 / 55.5 \\
    IE & Classical texts & Precision & 57.9 / 38.1 & 59.7 / 38.9 & 59.8 / 39.6 & 58.1 / 35.4 & 54.5 / 35.0 & 48.9 / 23.9 & 50.3 / 26.6 & 56.6 / 39.8 & 54.2 / 35.7 & \textbf{60.8} / \textbf{42.0} \\
    IE & Classical texts & Recall & 56.9 / 37.9 & 56.3 / 37.3 & 56.3 / 38.3 & 54.7 / 34.4 & 49.2 / 32.0 & 60.5 / 35.6 & 54.0 / 31.8 & 62.0 / 47.5 & 56.1 / 40.1 & \textbf{65.9} / \textbf{51.8} \\
    IE & Classical texts & F1 & 56.6 / 36.9 & 57.2 / 37.0 & 57.3 / 38.0 & 55.6 / 34.0 & 51.0 / 32.4 & 53.4 / 28.0 & 51.3 / 28.1 & 58.5 / 42.5 & 54.6 / 37.0 & \textbf{62.6} / \textbf{45.6} \\
    \midrule
    DTR & Syndrome & Precision & 7.7 / 0.6 & 8.4 / 1.4 & 7.6 / 1.2 & 7.5 / 1.0 & 6.9 / 1.4 & 3.4 / 0.3 & 3.3 / 1.4 & 10.6 / 1.9 & 9.1 / 2.3 & \textbf{12.3} / \textbf{2.5} \\
    DTR & Syndrome & Recall & 11.2 / 0.9 & 12.9 / 2.0 & 11.3 / 1.2 & 9.9 / 1.1 & 9.8 / 1.7 & 5.6 / 0.4 & 5.4 / 2.1 & 9.5 / 1.5 & 11.0 / \textbf{2.9} & \textbf{13.0} / 2.5 \\
    DTR & Syndrome & F1 & 8.8 / 0.7 & 9.8 / 1.6 & 8.8 / 1.2 & 8.2 / 1.0 & 7.8 / 1.5 & 4.1 / 0.3 & 3.9 / 1.6 & 9.5 / 1.6 & 9.5 / \textbf{2.5} & \textbf{12.1} / 2.4 \\
    DTR & Treatment & Precision & 11.5 / 1.3 & 13.0 / 0.8 & 11.0 / 0.8 & 12.0 / 0.8 & 11.6 / 1.3 & 13.4 / 1.3 & 6.5 / 0.8 & 14.7 / 2.3 & 12.9 / \textbf{2.6} & \textbf{16.5} / 2.4 \\
    DTR & Treatment & Recall & 14.4 / 1.5 & 15.4 / 1.1 & 14.3 / 1.1 & 14.7 / 1.0 & 14.4 / 1.7 & 15.7 / 1.5 & 7.8 / 0.9 & 14.2 / 2.6 & \textbf{17.6} / \textbf{3.8} & 15.6 / 2.0 \\
    DTR & Treatment & F1 & 12.3 / 1.3 & 13.6 / 0.9 & 12.0 / 0.9 & 12.8 / 0.9 & 12.4 / 1.5 & 12.9 / 1.2 & 6.7 / 0.8 & 13.9 / 2.4 & 14.3 / \textbf{3.0} & \textbf{15.5} / 2.1 \\
    DTR & Prescription & Precision & 31.0 / 17.0 & 32.4 / 18.1 & 31.1 / 17.2 & 31.7 / 17.1 & 30.4 / 16.7 & 27.6 / 16.1 & 24.8 / 15.7 & 35.5 / 24.0 & 34.3 / 21.9 & \textbf{36.6} / \textbf{24.3} \\
    DTR & Prescription & Recall & 29.3 / 16.9 & 26.0 / 14.9 & 29.6 / 17.2 & 26.4 / 14.7 & 25.0 / 14.0 & 30.8 / 19.0 & 21.2 / 14.0 & 38.2 / \textbf{27.3} & 36.2 / 24.8 & \textbf{38.9} / 27.2 \\
    DTR & Prescription & F1 & 29.5 / 16.4 & 28.3 / 15.8 & 29.8 / 16.7 & 28.2 / 15.2 & 26.8 / 14.7 & 28.5 / 16.9 & 22.2 / 14.2 & 36.2 / 24.8 & 34.5 / 22.6 & \textbf{37.0} / \textbf{24.9} \\
    DTR & Dosage & MAE & 4.4 / 4.7 & 4.4 / \textbf{4.0} & 4.2 / 4.2 & 4.3 / 4.1 & \textbf{4.1} / 4.1 & 5.3 / 5.3 & 4.5 / 4.3 & 4.6 / 4.7 & 4.4 / 4.6 & 4.3 / 4.4 \\
    DTR & Dosage & Cosine & 29.0 / 15.5 & 28.1 / 15.3 & 28.6 / 15.7 & 27.2 / 14.2 & 26.3 / 14.5 & 27.9 / 16.0 & 20.0 / 12.4 & 34.2 / 23.4 & 33.0 / 21.1 & \textbf{35.4} / \textbf{23.7} \\
    \midrule
    DTR-F1 & Syndrome & Precision & 16.5 / 7.8 & 17.3 / 9.2 & 16.5 / 8.3 & 17.4 / 8.2 & 15.5 / 8.1 & 10.1 / 4.7 & 7.7 / 3.4 & 23.8 / 11.6 & 20.8 / 12.5 & \textbf{25.1} / \textbf{12.8} \\
    DTR-F1 & Syndrome & Recall & 23.8 / 10.5 & \textbf{26.5} / 13.0 & 24.1 / 11.3 & 22.5 / 10.2 & 21.9 / 10.7 & 16.0 / 6.4 & 11.9 / 4.7 & 20.8 / 9.6 & 25.1 / \textbf{14.1} & 25.5 / 12.1 \\
    DTR-F1 & Syndrome & F1 & 18.7 / 8.7 & 20.1 / 10.4 & 18.8 / 9.2 & 18.8 / 8.8 & 17.5 / 8.8 & 11.9 / 5.3 & 8.8 / 3.8 & 21.1 / 9.8 & 21.6 / \textbf{12.7} & \textbf{24.2} / 11.9 \\
    DTR-F1 & Treatment & Precision & 18.0 / 6.7 & 20.1 / 7.7 & 16.8 / 6.2 & 18.8 / 6.1 & 17.4 / 6.2 & 17.2 / 4.0 & 10.3 / 3.8 & 22.6 / \textbf{8.7} & 19.2 / 7.3 & \textbf{25.1} / 8.5 \\
    DTR-F1 & Treatment & Recall & 23.0 / 9.0 & 23.9 / 10.2 & 22.4 / 8.8 & 23.4 / 8.5 & 21.8 / 8.6 & 22.6 / 7.1 & 13.4 / 5.4 & 21.8 / 9.1 & \textbf{26.4} / \textbf{11.0} & 23.7 / 8.4 \\
    DTR-F1 & Treatment & F1 & 19.5 / 7.5 & 21.1 / \textbf{8.6} & 18.6 / 7.1 & 20.2 / 6.9 & 18.8 / 7.1 & 17.4 / 4.8 & 11.1 / 4.3 & 21.5 / \textbf{8.6} & 21.4 / \textbf{8.6} & \textbf{23.6} / 8.2 \\
    DTR-F1 & Prescription & Precision & 30.9 / 17.0 & 32.4 / 18.1 & 30.9 / 17.1 & 31.7 / 17.2 & 30.4 / 16.8 & 27.7 / 16.3 & 25.4 / 15.7 & 35.5 / 24.0 & 33.9 / 21.9 & \textbf{36.7} / \textbf{24.3} \\
    DTR-F1 & Prescription & Recall & 29.6 / 17.0 & 26.0 / 14.9 & 29.9 / 17.4 & 26.5 / 14.8 & 25.1 / 14.1 & 31.1 / 19.2 & 21.9 / 14.2 & 38.3 / \textbf{27.2} & 36.8 / 25.2 & \textbf{38.9} / 27.1 \\
    DTR-F1 & Prescription & F1 & 29.6 / 16.5 & 28.3 / 15.8 & 29.8 / 16.8 & 28.3 / 15.3 & 26.9 / 14.8 & 28.7 / 17.1 & 22.9 / 14.4 & 36.2 / 24.8 & 34.6 / 22.8 & \textbf{37.1} / \textbf{25.0} \\
    DTR-F1 & Dosage & MAE & 4.4 / 4.7 & 4.4 / \textbf{4.0} & 4.2 / 4.3 & 4.3 / 4.1 & \textbf{4.1} / 4.1 & 5.3 / 5.4 & 4.6 / 4.4 & 4.6 / 4.7 & 4.4 / 4.7 & 4.3 / 4.4 \\
    DTR-F1 & Dosage & Cosine & 29.2 / 15.6 & 28.2 / 15.3 & 28.7 / 15.8 & 27.4 / 14.5 & 26.3 / 14.6 & 28.1 / 16.2 & 20.8 / 12.5 & 34.3 / 23.5 & 33.3 / 21.4 & \textbf{35.6} / \textbf{23.8} \\
    \midrule
    DR & Syndrome & Accuracy & 86.0 / 39.3 & 84.1 / 29.5 & 84.2 / 34.5 & 82.8 / 29.0 & 77.3 / 25.3 & 80.7 / 26.0 & 70.3 / 30.3 & \textbf{91.4} / \textbf{62.3} & 87.0 / 46.8 & 90.8 / 57.0 \\
    DR & Treatment & Accuracy & 80.0 / 22.3 & 80.4 / 23.8 & 78.4 / 19.0 & 78.6 / 20.3 & 74.8 / 24.0 & 78.1 / 26.0 & 68.2 / 33.5 & \textbf{88.1} / \textbf{53.5} & 84.1 / 40.0 & 86.9 / 47.0 \\
    DR & Prescription & Accuracy & 87.5 / 45.3 & 86.7 / 40.5 & 87.4 / 43.3 & 84.9 / 35.3 & 82.4 / 36.0 & 84.4 / 39.5 & 64.5 / 25.5 & 91.5 / 61.8 & 91.9 / 63.0 & \textbf{93.4} / \textbf{69.3} \\
    \midrule
    \multicolumn{2}{c}{\textbf{Overall Score}} &  & 62.3 / 34.5 & 60.2 / 30.8 & 60.8 / 31.4 & 59.0 / 28.8 & 55.9 / 27.6 & 56.7 / 27.2 & 46.3 / 24.1 & 63.4 / 40.4 & 63.7 / 39.1 & \textbf{67.1} / \textbf{45.3} \\
    \bottomrule
    \end{tabular}%
    }
\end{table}

\clearpage
\section{Information Extraction Performance}

\begin{figure}[pos=!hbp]
    \centering
    \includegraphics[width=\linewidth]{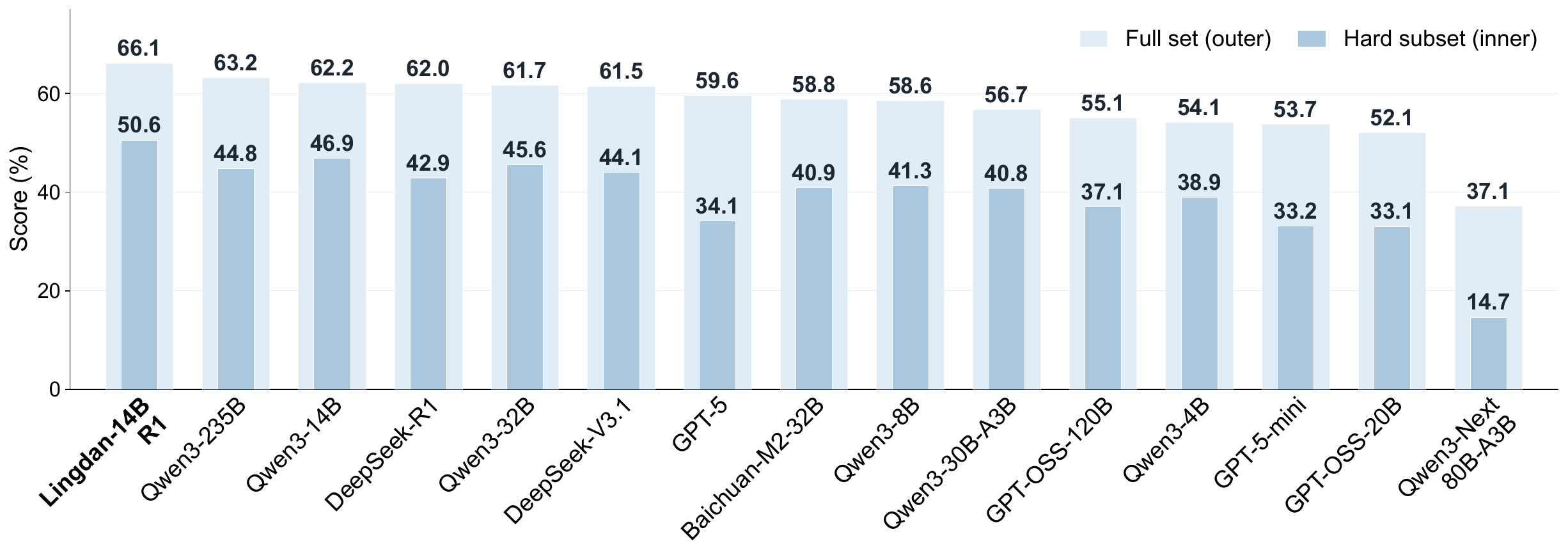}
    \caption{Information extraction on LingLan. Scores are the equal-weight mean of clinical-record and classical-text extraction F1. Wide light-blue bars show full-benchmark scores, and narrow dark-blue bars show LingLan-Hard scores.}
    \label{fig:linglan_information_extraction}
\end{figure}

\clearpage
\section{Retrospective Clinical Prescription Evaluation}

\begin{figure}[pos=!hbp]
    \centering
    \includegraphics[width=\linewidth]{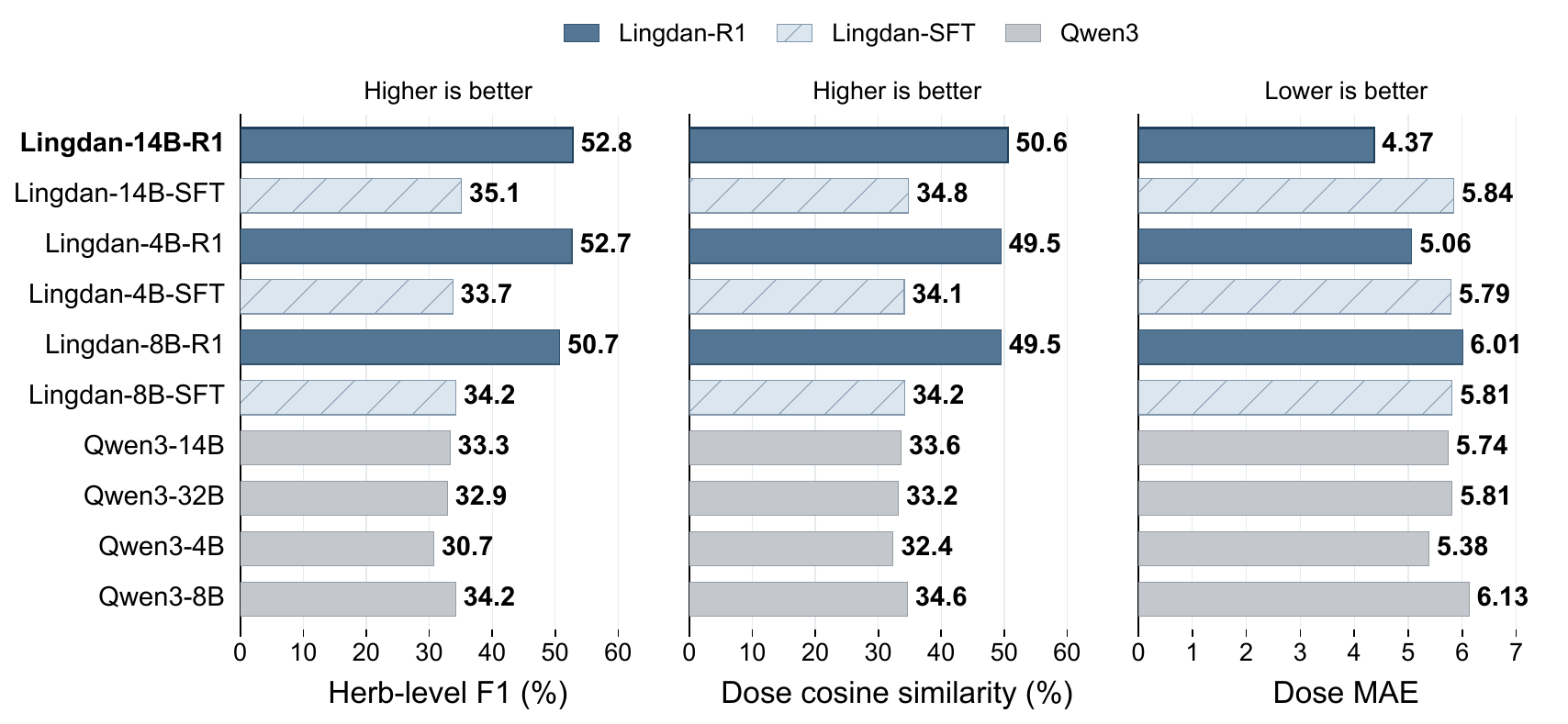}
    \caption{Prescription recommendation on the private spleen-stomach disease test set. From left to right, the plots show herb-level F1, dose cosine similarity, and dose MAE. Dark-blue bars represent Lingdan-R1 models, light-blue hatched bars represent Lingdan-SFT models, and light-grey bars represent Qwen3 reference models. Higher F1 and cosine similarity indicate better agreement; lower MAE indicates lower absolute dosage error.}
    \label{fig:pwb_prescription}
\end{figure}

\clearpage
\section{Reward Optimization under the 10-Physician Configuration}

\begin{figure}[pos=!hbp]
    \centering
    \includegraphics[width=\linewidth]{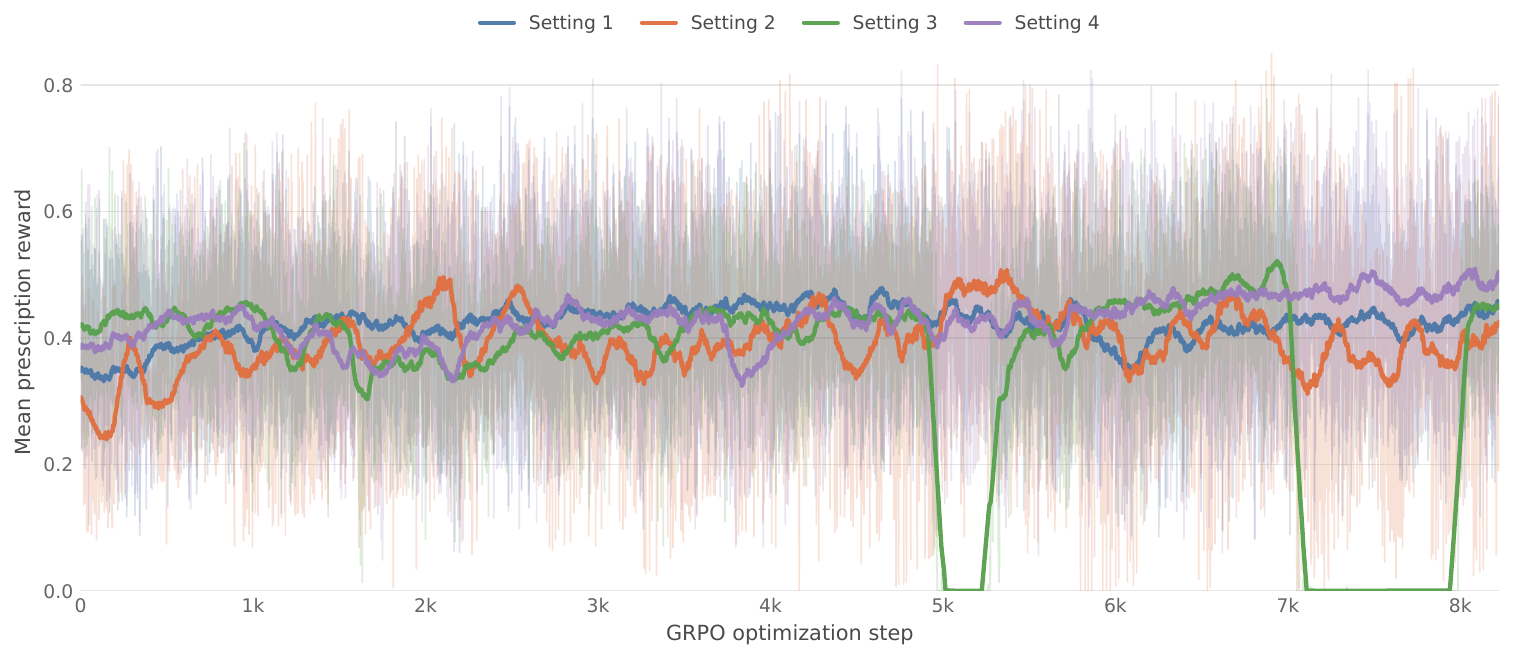}
    \caption{GRPO prescription reward trajectories for four Lingdan-4B development runs using the 10-physician source pool under different hyperparameter settings. Faint curves show raw rewards, and solid curves show moving averages calculated over a fixed 100-step window.}
    \label{fig:appendix_top10_runs}
\end{figure}

To further investigate the optimization behavior of the 10-physician configuration, we conducted four Lingdan-4B development runs under different hyperparameter settings. As shown in Figure~\ref{fig:appendix_top10_runs}, the reward trajectories exhibited considerable fluctuations and limited long-term improvement. Although transient increases were observed under some settings, these gains were not consistently maintained throughout training.
Across the four runs, reward improvement remained limited and was not sustained throughout training.
Together with the physician-source comparison in Figure~\ref{fig:physician_source_dynamics}, these results further demonstrate the influence of physician-source heterogeneity on prescription-centered reinforcement learning.

\end{appendices}
\end{document}